\documentclass[preprint,11pt,authoryear]{elsarticle}

\usepackage{setspace}
\usepackage{amsmath}
\usepackage{booktabs}
\usepackage{amsfonts}
\usepackage{mathtools}
\usepackage{amssymb}
\usepackage{mathrsfs}
\usepackage{makecell}
\usepackage[a4paper, total={6.5in, 9.5in}]{geometry}
\usepackage{multirow}
\usepackage{lineno}
\usepackage[ruled,vlined]{algorithm2e}
\usepackage[
    colorlinks=true,
    linkcolor=blue,
    citecolor=blue,
    urlcolor=blue
]{hyperref}
\usepackage{enumitem} 

\makeatletter
\AtBeginDocument{%
    \onehalfspacing
    \global\let\@blstr\baselinestretch
}
\makeatother

\newtheorem{theorem}{Theorem}

\newtheorem{corollary}{Corollary}

\newdefinition{definition}{Definition}
\newdefinition{remark}{Remark}
\newdefinition{observation}{Observation}

\newtheorem{example}{Example}
\newtheorem{assumption}{Assumption}
\newtheorem{proposition}{Proposition}

\newcommand{\proofofref}{}
\newproof{proofofaux}{Proof of \proofofref}

\newenvironment{proofof}[2]
{%
    \renewcommand{\proofofref}{#1~\ref{#2}}%
    \begin{proofofaux}%
}
{%
    \end{proofofaux}%
}

\begin{document}
\let\P\relax
\let\L\relax
\newcommand{\P}{\mathbb{P}}
\newcommand{\L}{\mathscr{L}}
\newcommand{\Q}{\mathbb{Q}}
\newcommand{\R}{\mathbb{R}}
\newcommand{\E}{\mathbb{E}}
\newcommand{\Z}{\mathcal{Z}}
\newcommand{\Y}{\mathcal{Y}}
\newcommand{\bz}{\boldsymbol{z}}
\newcommand{\set}[1]{ \left\{ #1 \right\} }
\newcommand{\bb}[1]{\boldsymbol{#1}}
\newcommand{\lra}[1]{\left(#1\right)}
\newcommand{\lrb}[1]{\left[#1\right]}
\newcommand{\lrc}[1]{\left\{#1\right\}}
\newcommand{\txi}{\tilde{\xi}}
\newcommand{\bxi}{\boldsymbol{\xi}}
\newcommand{\argmin}{\mathop{\arg\min}}
\newcommand{\argmax}{\mathop{\arg\max}}
\renewcommand{\bibfont}{\small}
\begin{frontmatter}



\title{Data-Driven Maintenance Optimization:\\Aligning Prognostic Learning with Decision Objective} 

\author[label1]{Zhuojun Xie}
\author[label1]{Adam Abdin}
\author[label1]{Yiping Fang \corref{cor1}}
\cortext[cor1]{Corresponding author.}
\affiliation[label1]{organization={Laboratoire Génie Industriel, CentraleSupélec, Paris-Saclay University},
                country={France}}

\begin{abstract}
Estimation error in machine learning prognostic models can propagate into poor maintenance decisions. We study a data-driven predictive maintenance problem coupling health-condition estimation with a joint ordering-replacement model. We show that the conventional Estimate-Then-Optimize (ETO) approach yields inconsistent, suboptimal decisions for a broad class of such problems. To bridge the gap, we propose an Integrated Estimate-Optimize (IEO) framework that directly targets maintenance cost as the prognostic learning objective, bypassing surrogate statistical accuracy metrics such as Kullback--Leibler divergence or mean squared error. Theoretically, we prove out-of-sample performance guarantees of the IEO framework under certain conditions, including i.i.d. and sliding-window data-generating processes. 
Computationally, we develop a training procedure combining Monte Carlo dropout, central-difference gradient approximation, and an ETO warm-start to handle discrete maintenance decisions and nonlinear costs. Experiments on the four CMAPSS subsets demonstrate that, under the nominal setting, IEO reduces mean maintenance cost by $14.8$--$22.7\%$ relative to ETO and by $7.2$--$11.8\%$ relative to a decision-oriented baseline, while also reducing failure rates and cost variation. These improvements are obtained despite higher prediction errors. Increasing the ratio of unplanned-downtime costs to planned-downtime costs makes this tradeoff more pronounced: IEO shifts predictions further toward RUL underestimation, accompanied by greater cost savings and fewer failures. These findings support aligning prognostic learning with maintenance objectives when imperfect predictions must guide decisions under costly unplanned downtime.
\end{abstract}


\end{frontmatter}

\section{Introduction}\label{sec:1}
\subsection{Background}
Preventive maintenance strategies have become particularly prevalent in asset-intensive industries due to their effectiveness in improving operational efficiency and asset management. For instance, organizations such as airlines~\citep{Lagos-2020-TS} and railways~\citep{Wagenaar-2017-TS} increasingly adopt preventive maintenance to enhance efficiency, minimize unplanned downtime, and streamline maintenance management~\citep{DeJonge-2020-EJOR}. Condition-Based Maintenance (CBM) has been widely adopted across industries to schedule preventive maintenance based on asset health, thereby enabling more effective resource allocation, reducing unexpected downtime, and improving operational reliability~\citep{Arts-2016-OR, Mookerjee-2023-POM}.

While CBM and Predictive Maintenance (PdM) are sometimes used synonymously, we distinguish PdM from CBM by its reliance on forecasting the future degradation state of the system, rather than assessing its current condition~\citep{Van-2023-IJPR}. 
A range of CBM approaches assume a fixed model class for asset lifetime distributions or degradation processes, such as Poisson processes~\citep{Drent-2023-MSOM}, non-homogeneous Poisson processes~\citep{Lee-2016-EJOR}, and Gamma processes \citep{Khatab-2019-IJPR}, while allowing the associated model parameters to vary. However, invariant statistical models hardly adapt to varying operational conditions, which can lead to suboptimal maintenance schedules, increased downtime, and frequent service interruptions. In contrast, PdM leverages real-time sensor data and advanced data analytics for health-condition prediction, and integrates seamlessly with a variety of maintenance decision-making models.

Interest in data-driven PdM strategies has been growing, fueled by real-time sensor monitoring techniques which circumvent the obstacle of direct condition observation~\citep{Randall-2025-EJOR, Okumucsouglu-2024-COMPUTING}. Most data-driven PdM approaches utilize real-time sensor data to estimate the health-condition of assets in terms of Remaining Useful Life (RUL) and its distribution, Remaining Life Distribution (RLD). This real-time predictive process, termed prognostics~\citep{Lei-2018-MSSP}, guides timely maintenance decision-making. Machine Learning (ML) models have become key enablers for data-driven PdM for their well-known ability to capture nonlinear and time-varying dependencies in data. Nonetheless, model misspecification\footnote{We formally define model misspecification in Definition~\ref{def:mis}.}~\citep{Lv-2014-JRS} is a fundamental and, in practice, unavoidable challenge for ML-based PdM, because flexible models, which have complex architectures and large numbers of parameters, typically demand large samples for model validation, training, and generalization~\citep{Bartlett-2021-Deeplearning}. In prognostics and health management, however, this requirement is seldom satisfied, as run-to-failure experiments, which are necessary to obtain labeled data for learning algorithms, are costly and time-consuming~\citep{hakami-2024-SP}. Consequently, the scarcity of training data inevitably leads to some degree of model misspecification in practical PdM applications.

A critical question therefore arises in data-driven PdM~\citep{Gebraeel-2023-JDMD}: \textit{how can inherently imperfect predictive models still guide effective maintenance planning?} A recent line of research in data-driven optimization has highlighted the advantages of decision-aware learning for decision-making in the presence of contextual information and model misspecification~\citep{DFL}. In contrast to the prevalent estimate-then-optimize (ETO) approach, which prioritizes estimation accuracy, decision-aware learning produces decision-centric estimates by explicitly emphasizing the downstream decision objective throughout training. It has been shown that the ETO approach can yield relatively better predictive accuracy but inferior decision quality due to the misalignment between the learning objective and the downstream decision problem~\citep{SPO}.

Motivated by these insights, we propose a data-driven maintenance framework that integrates the prognostic model and the optimization model into an end-to-end policy. In particular, we focus on a single-unit joint ordering-replacement model that unifies inventory, downtime, and replacement decisions~\citep{Nguyen-2019-RESS, moghaddass-2018-AOR}, capturing their complex interaction under uncertainty. To the best of our knowledge, this is the first study in data-driven PdM to align prognostic model learning and maintenance optimization with theoretical guarantees. We further provide practical insights into the tradeoff between estimation accuracy and decision quality under different levels of aversion to operational risk, which in this work is represented by unplanned downtime.

\subsection{Related work}
This section reviews three streams of literature closely related to the data-driven decision-oriented PdM framework: (i) recent empirical studies in data-driven PdM, (ii) limited work on decision-oriented data-driven maintenance, and (iii) advances in data-driven contextual optimization.

\subsubsection{Data-driven predictive maintenance}
The main line of research in data-driven PdM predominantly follows the ETO framework. Under this framework, probabilistic prognostic models have recently gained popularity in data-driven PdM due to their ability to quantify uncertainty. A key advantage of probabilistic models in maintenance optimization is their compatibility with various maintenance strategies that rely on RLD, whereas deterministic models are typically incorporated into simple threshold-based policies. \cite{Lee-2023-RESS} employed a convolutional neural network combined with dropout to generate estimates for RLDs. They then trained a deep reinforcement learning algorithm for sequential maintenance planning using these estimates. The same neural architecture was adopted by \cite{mitici-2023-RESS}, who combined it with single- and multi-component replacement models for aero-engine maintenance management.
Other notable works include the approach by \cite{Nguyen-2019-RESS}, who developed a Long Short-Term Memory (LSTM) classifier to estimate failure probabilities, integrating these predictions into a joint ordering-replacement model. 
These studies predominantly adopt the ETO framework, optimizing ML models primarily for estimation accuracy, with limited attention to their impact on maintenance decisions. Moreover, they have not explicitly investigated the link between estimation accuracy and decision quality, leaving the previous critical question unanswered. To address this gap, we fine-tune a prognostic model directly with respect to the downstream maintenance cost, thereby making the tradeoff between estimation accuracy and maintenance cost explicit.

\subsubsection{Decision-oriented prognostics and maintenance}
A few recent studies have emphasized that prognostic models should be evaluated in light of their impact on downstream health-management tasks. \cite{Cai-2023-EJOR} developed a nonparametric MLE method to estimate failure probabilities from data and integrated these estimates into a threshold policy. Although the threshold was determined based on maintenance costs, it was chosen separately from the estimation process. This separation between estimation and optimization is persistent in the literature. \cite{Sun-2023-POMS} considered a joint production-maintenance problem and addressed the degradation uncertainty by robust optimization, whereas degradation model parameters were estimated prior to the decision-making stage. Similarly, \cite{Okumucsouglu-2024-COMPUTING} combined sensor-driven RLDs into a two-stage stochastic maintenance–operation framework, yet the distributions were treated as fixed inputs, thereby preserving the open-loop separation between estimation and optimization. The recent work by~\citet{Kamariotis-2024-RESS} proposed a decision-oriented metric to compare prognostic algorithms and guide hyperparameter tuning. However, their hyperparameter tuning heuristics offer no guarantees regarding decision quality.~\cite{Van-2025-SSRN} proposed a decision rule approach that directly maps sensor data into binary variables for all feasible decisions without modeling the uncertainty. The parameters in the decision rule were optimized for maintenance cost in a decision-oriented manner. Nevertheless, the decision rule was limited to one decision variable. By contrast, the proposed framework explicitly quantifies uncertainty in RUL estimates and integrates an optimization-based model for maintenance decision-making under heterogeneous decision variables.

\subsubsection{Data-driven contextual optimization}
Data-driven PdM can be viewed as an instance of contextual optimization under uncertainty, where sensor readings serve as contextual information observed prior to decision-making, and the RUL is the random outcome realized after decisions are made. Two primary methodological paradigms exist for data-driven contextual optimization~\citep{Sadana-2025-EJOR}: \textit{decision rule optimization} and \textit{learning-and-optimization}. Decision rule approaches directly map observed contextual data into feasible decisions by linear/nonlinear transformations without explicitly modeling uncertainty~\citep{Ban-2019-OR, Van-2025-SSRN}. Learning-and-optimization approaches instead model conditional uncertainty (either as point predictions or distribution estimates), and integrate the uncertainty into downstream optimization. Nonparametric methods, such as weighted sample average approximation~\citep{Bertsimas-2020-MS}, estimate conditional distributions directly from historical data but ignore decision-making signals and may perform poorly with limited samples. Moreover, nonparametric approaches are known to struggle with high-dimensional features~\citep{Stone-1980-AS}, and therefore have not been widely considered in the maintenance literature. In this work, we adopt parametric models, which better accommodate such high-dimensional and dependent features, e.g., sensor observations. Recent research has further emphasized integrating parametric models with optimization objectives in a decision-oriented manner. Seminal studies by~\cite{SPO} and~\cite{DFL}, along with subsequent work, have developed numerous techniques to smooth this integration mainly for deterministic predictive models and linear programs. Addressing this gap, we introduce the first Integrated Estimate-Optimize (IEO) framework for data-driven PdM with theoretical guarantees, integrating a parametric prognostic model and a combinatorial optimization model with a nonlinear, nonconvex objective in the decision variables. On the computational side, we propose an efficient training strategy that combines Monte Carlo (MC) dropout and central-difference gradient approximation for generic neural networks to advance the application of IEO to data-driven maintenance problems.

\subsection{Contributions}
In summary, the contributions of this work are as follows:
\begin{enumerate}
    \item We develop the first IEO framework for data-driven PdM with provable finite-sample guarantees, integrating probabilistic prognostics with combinatorial maintenance optimization under nonlinear costs. We identify a structural limitation of ETO that motivates this integration: arbitrarily close RLD estimates can induce distinct maintenance decisions, and estimation accuracy alone does not ensure decision quality.
    
    \item We establish the consistency of empirical decision-risk minimization and derive guarantees under independent and sliding-window-dependent sampling. In addition, we verify the finite Natarajan dimension of our policy class.
    
    \item We develop a practical IEO learning algorithm combining MC dropout, central-difference gradient approximation, and an ETO warm-start to train prognostic models using maintenance costs despite discrete decisions and nonlinear objectives. The surrogate gradient is justified through a locally averaged decision loss.

    \item Across the four CMAPSS subsets, IEO reduces mean maintenance cost by $14.8$--$22.7\%$ relative to ETO and $7.2$--$11.8\%$ relative to decision-oriented tuning under the nominal setting, despite larger prediction errors. Further experiments show that higher penalties for unplanned downtime generally strengthen these benefits, clarifying when aligning prognostic learning with maintenance costs is most valuable.
\end{enumerate}
\section{Problem Setting}\label{sec:2}
We study a sensor-driven joint ordering-replacement problem in which the decision-maker uses the latest
sensor observations of an asset to make maintenance decisions under RUL uncertainty. We focus on
a setting where decisions must be tailored repeatedly across assets of the same type but with different
sensor observations, reflecting heterogeneous degradation states. Formally, the decision-maker seeks a policy $\pi$ that minimizes
\begin{equation}
    \E_{\P}\lrb{c\lra{\pi(X); \bxi}}, \label{policyobj}
\end{equation}
where $X$ is the random sensor observation, $\bxi$ is the random RUL whose value is revealed only after the decision is made,  $\P$ is the joint distribution of $(X,\bxi)$, $\pi$ maps sensor observations to feasible decisions, and $c$ is the maintenance objective
function. The expectation is taken over the joint distribution $\P$, so a single policy $\pi$ is applied across realizations of $X$. Throughout, we assume the decision-maker has no access to the true joint distribution $\P$; only a dataset $\mathcal{D}$ of paired sensor data and RUL labels, collected from run-to-failure experiments, is available.

The remainder of this section is organized as follows. Section~\ref{sec:2_1} specifies the decision variables and
the objective function $c$ of the ordering-replacement problem; Section~\ref{sec:2_2} formalizes the notation and the
RUL modeling; and Section~\ref{sec:2_3} constructs a prognosis-then-maintenance policy.


\subsection{Decision variables and objective}\label{sec:2_1}
Let $\bz=(z_o,z_d,z_r)$ collect the ordering time $z_o$, the planned decommissioning time $z_d$, and the scheduled replacement time $z_r$. Each variable measures the number of unit cycles from the current decision epoch to the corresponding action. We distinguish decommissioning $z_d$ from replacement $z_r$ because an operator may take the asset out of service before a scheduled maintenance visit to avoid failure. The feasible sets $\mathcal{Z}_o \ni z_o$, $\mathcal{Z}_d \ni z_d$, and $\mathcal{Z}_r \ni z_r$ are finite sets of nonnegative integers. For concreteness, consider
\begin{equation}
    \mathcal{Z}_o = \{0, 1, \dots, \bar{Z}\}, \quad \mathcal{Z}_d = \{0, 1, \dots, \bar{Z}\}, \quad \mathcal{Z}_r = \{\Delta_r, \dots, N\Delta_r\}, \notag
\end{equation}
where $\bar{Z},\, N$ and $\Delta_r$ are positive integers with $\Delta_r \geq 2$. The arithmetic progressions are chosen for generality; other choices are admissible and affect neither our algorithm nor our analysis. Ordering and decommissioning can be scheduled after any cycle within their respective horizons, whereas replacement is scheduled according to maintenance-crew availability. Other finite discrete schedules can be accommodated within the same framework. We define the aggregated feasible set as
\begin{equation*}
\mathcal{Z} \coloneqq \lrc{ (z_o, z_d,z_r) \mid z_o \in \mathcal{Z}_o, z_d \in \mathcal{Z}_d, z_r \in \mathcal{Z}_r, z_d \leq z_r, z_o \leq z_r},
\end{equation*}
and therefore a policy $\pi$ maps the sensor data $X$ to a decision $\boldsymbol{z} \in \mathcal{Z}$.

Let $\boldsymbol{\tau}$ denote the stochastic spare-part lead time. The spare part arrives at $z_o+\boldsymbol{\tau}$, and the actual replacement time is the derived random variable:
\begin{equation*}
    T_r\coloneqq\max\{z_r,z_o+\boldsymbol{\tau}\}.
\end{equation*}
Thus, replacement occurs at the scheduled time if the spare part is available; otherwise, it is delayed until the part arrives. This formulation assumes that replacement can proceed upon arrival in the latter case. For a realized RUL $\xi$, we define the maintenance cost as
\begin{align}
    c(\bz;\xi) \coloneqq \mathbb E_{\boldsymbol{\tau}}\Big[ 
      &\underbrace{ c_h\bigl[T_r-(z_o+\boldsymbol{\tau})\bigr] }_{(a)} \notag \\
    + &\underbrace{ c_p\bigl[ [\xi-z_d]^+ +[T_r-\max\{\xi,z_d\}]^+ \bigr] }_{(b)} \notag \\
    + &\underbrace{ c_f[T_r-\xi]^+\mathbf 1[\xi<z_d] }_{(c)}
    \Big],\label{obj}
\end{align}
where $c_h$, $c_p$, and $c_f$ are the holding, planned-downtime, and unplanned-downtime costs per unit cycle, respectively; $[a]^+=\max\{a,0\}$; and $\mathbf 1[\cdot]$ is the indicator function. The expectation is taken over the distribution of $\boldsymbol{\tau}$. Since our focus is on RUL uncertainty, we assume that the lead-time distribution is known, allowing $c(\boldsymbol z;\xi)$ to be evaluated for each pair $(\boldsymbol z,\xi)$. The cost function in Equation~\eqref{obj} captures three economic impacts of a given decision $\boldsymbol{z}$: 
\begin{enumerate}[label=(\alph*)]
    \item the holding cost incurred between spare-part arrival at $z_o+\boldsymbol{\tau}$ and actual replacement at $T_r$;
    \item two opportunity-cost penalties: $[\xi-z_d]^+$ captures the useful life forfeited by decommissioning the asset before failure, while $[T_r-\max\{\xi,z_d\}]^+$ captures the production disruption when replacement is delayed beyond both decommissioning and failure;
    \item the unplanned-downtime cost incurred between failure at $\xi$ and actual replacement at $T_r$, applicable only when failure precedes planned decommissioning.
\end{enumerate}
Terms (a) and (c) follow the standard joint ordering-replacement formulation~\citep{Elwany-2008-IIET, Kamariotis-2024-RESS}. We introduce the two opportunity-cost terms in (b) to account explicitly for the distinct economic consequences of premature decommissioning and delayed replacement.
\subsection{RUL modeling}\label{sec:2_2}
The objective in Equation~\eqref{obj} presumes a known RUL, but in sensor-driven maintenance the RUL is typically uncertain at the decision stage. The decision-maker must therefore rely on data analytics to model RUL uncertainty for decision support. This section introduces a discrete model of the RLD that represents this uncertainty.

Consider an individual asset subject to stochastic deterioration under normal operating conditions, and let the RLD serve as its health-condition indicator. We model the RUL by a discrete random variable $\xi$ with finite support $\Xi \coloneqq \{\xi^1,\ldots,\xi^H\}\subset\mathbb{Z}_{\geq 0}$, reflecting the industrial practice in which asset lifetimes are recorded on a discrete timeline rather than in continuous clock time; for instance, in flight cycles for aircraft engines~\citep{CMAPSS} or charge/discharge cycles for batteries~\citep{Battery}. Assuming a unit cycle can be defined for the asset, each support point $\xi^h=h-1$ is associated with an RUL interval $I_h$. For $h=1,\ldots,H-1$, let $I_h=[h-1,h)$, while $I_H=[H-1,\infty)$ absorbs the upper tail. The horizon length $H$ is set by the modeler according to the asset characteristics and the decision structure. 

The dataset $\mathcal{D}$ may contain relatively few samples, owing to the limited number of run-to-failure experiments, and the data are typically high-dimensional and heterogeneous. The sensor observation, known to correlate with RUL, is a random variable $X$ with realization $x \in \mathcal{X} \subset \R^p$. Let $\P_X$ and $\P_{\bxi}$ denote the marginals of $\big(X,\bxi\big)$; the conditional RLD given $X$, denoted by $\P_{\bxi \mid X}$, is hereafter referred to as the conditional RLD. Let $\mathcal{P}_{\Xi} \coloneqq \{ P \in \R^{H}_{+}: \sum_{\xi \in \Xi} P(\xi) = 1 \}$ be the set of all probability distributions on $\Xi$, and let $\hat{P}_{\bxi \mid X} \in \mathcal{P}_\Xi$ denote an estimate of the conditional $\P_{\bxi\mid X}$.
\subsection{Prognosis-then-maintenance policy}\label{sec:2_3}
If the conditional RLD $\P_{\bxi \mid X}$ were known for every $X=x$, the decision-maker could draw on a range of tools to identify an optimal decision with respect to the RLD. We adopt the Contextual Stochastic Optimization (CSO) framework~\citep{Bertsimas-2020-MS}, which tailors the decision to each realization of $X$. Under CSO, an optimization model $g:\mathcal{P}_\Xi \rightarrow \mathcal{Z}$ prescribes a decision for the conditional $\mathbb{P}_{\bxi\mid X}$:
\begin{equation}
    \textbf{(CSO)} \quad g(\P_{\bxi \mid X}) \coloneqq \argmin_{\boldsymbol{z} \in \mathcal{Z}} \, \mathbb{E}_{\P_{\bxi\mid X}}\lrb{c(\boldsymbol{z};\bxi)}. \label{CSO}
\end{equation}
Because the feasible set $\mathcal{Z}$ is discrete and the \textbf{CSO} objective is affine with respect to the conditional distribution $\P_{\bxi \mid X}$, we precompute the objective values $c(\bz; \xi)$ for all $\bz \in \mathcal{Z}$ and $\xi \in \Xi$. For each candidate conditional distribution $\P_{\bxi \mid X}$, we then evaluate $\mathbb{E}_{\P_{\bxi \mid X}}\!\left[c(\bz; \bxi)\right]$ and select the decision $\bz$ that minimizes this conditional expected objective.
\begin{remark}[CSO versus Renewal-Reward Model]
    The \textbf{CSO} formulation in Equation~\eqref{CSO} is context-specific and differs from the renewal-reward model. For a realized sensor observation $X=x$, the conditional RLD $\P_{\bxi\mid X=x}$ characterizes the RUL uncertainty of the asset at the current decision epoch, and the prescribed decision $g(\P_{\bxi\mid X=x})$ is therefore evaluated over the ensuing maintenance episode. We do not assume that, after replacement, the system regenerates under the same context $x$ and repeatedly follows the same conditional RLD. Instead, subsequent decisions are triggered by newly observed sensor contexts, which may represent different degradation states and operating histories. Accordingly, long-run performance for multiple assets is captured at the policy level through \(\mathbb{E}_{\P}[c(\pi(X);\bxi)]\) in Equation~\eqref{policyobj}, which aggregates costs across heterogeneous contexts, rather than through a renewal-reward ratio defined for repeated assets that share the same conditional RLD.
\end{remark}

A crucial gap concerns the dependence of $g$ on its input, the conditional RLD: in a data-driven setting the conditional can only be estimated. The prevailing practice in data-driven PdM fits a parametric model to estimate the conditional and then feeds these estimates to the downstream optimization model. This approach is known as estimate-then-optimize, or \textit{prognosis-then-maintenance} in the context of PdM.

Let $\mathcal{M} \coloneqq \lrc{m(\cdot;\omega):\mathcal{X} \rightarrow \mathcal{P}_\Xi \mid \omega \in \Omega }$ be a parametric prognostic model class, determined by the architecture of model $m$ and a parameter space $\Omega \subset \R^q$. Composing the \textbf{CSO} model with the predictive model yields the prognosis-then-maintenance policy class in Definition~\ref{def:decision-rule}.
\begin{definition}\label{def:decision-rule}
    A policy class 
    $$\Pi \coloneqq \big\{\pi_\omega(\cdot) \coloneqq g\lra{m(\cdot;\omega)}:\mathcal{X} \rightarrow \mathcal{Z} \mid \omega \in \Omega\big\}$$ 
    is induced by the optimization model $g$ in Equation~\eqref{CSO} and a parametric prognostic model class $\mathcal{M}$.
\end{definition}

Hereafter, we approximate the optimal policy for Problem~\eqref{policyobj} by a policy $\pi_\omega \in \Pi$, thereby reducing the search over policies to a search over the parameter $\omega$. The parameter $\omega$ is therefore central, as it shapes the conditional estimates. In the conventional ETO approach, $\omega$ is learned by minimizing the in-sample estimation error and is then applied to unseen $X$ to estimate the conditionals, which in turn serve as input to the policy. Section~\ref{sec:3} exposes the shortcomings of this approach in our setting.
\section{Consistency Analysis}\label{sec:3}
In this section, we briefly revisit the ETO approach and demonstrate, through empirical and theoretical evidence, its potential to produce inconsistent decisions. 
We then introduce the notion of \textit{consistency}, which characterizes the relationship between RLD estimation and maintenance optimization in PdM.
Finally, we establish a finite-sample guarantee, providing theoretical support for the IEO approach that is consistent with the decision-making process. 
Overall, the analysis explains why ETO can produce inconsistent maintenance decisions and motivates the IEO framework.
\subsection{Estimate-then-optimize}\label{sec:ETO}

Denote the dataset by $\mathcal{D} \coloneqq \lrc{\lra{x_i, \xi_i}}_{i=1}^{n}$ containing $n$ pairs of sensor data $x_i$ and RUL labels $\xi_i$. The conventional ETO approach often adopts a prognostic model $m\lra{\cdot;\omega}$ to estimate the conditionals, and learns the model parameter $\omega$ by minimizing the empirical estimation error over the dataset $\mathcal{D}$ defined in Equation~\eqref{functional-ETO}:
\begin{equation}
    \hat{\omega}_{eto} \coloneqq 
    \argmin_{\omega \in \Omega}\frac{1}{n}\sum_{i = 1}^{n} \ell_{eto}\lra{m(x_i;\omega),\xi_i} + \lambda \Vert \omega \Vert
    \label{functional-ETO},
\end{equation}
where $m$ is a prognostic model from a prespecified model class $\mathcal{M}$ with $\Omega$ being a parameter space as in Definition~\ref{def:decision-rule}, $\ell_{eto}$ a loss function, and $\lambda \geq 0$ the strength of regularization. Specifically, the Kullback--Leibler divergence (KL-div) $\mathbb{D}_{\mathrm{KL}}$ defined in Equation~\eqref{KLdiv} is often adopted as the loss function for probabilistic estimation:
\begin{equation} \label{KLdiv}
    \mathbb{D}_{\mathrm{KL}}(P \Vert Q) \coloneqq \sum_{\xi \in \Xi} P(\xi)\log \frac{P(\xi)}{Q(\xi)}, \quad \forall P, Q \in \mathcal{P}_\Xi.
\end{equation}
For a dataset that contains only point realizations of $\bxi$, the negative log-likelihood (if a specific density is assumed) or the Mean Squared Error (MSE) is often adopted as the loss function $\ell_{eto}$ for model training.

Once the (near-)optimal model parameter $\hat{\omega}_{eto}$ has been obtained, the ETO approach leverages the learned parameter in the policy $\pi_{\hat{\omega}_{eto}}$ to estimate the RLD for an unseen instance $x_0$. The estimate $m(x_0;\hat{\omega}_{eto})$ is then passed through the decision model $g$ to prescribe the ordering and replacement decisions. Recent theoretical developments highlight important limitations of the ETO approach, particularly for decision-making under considerable model misspecification~\citep{Elmachtoub-2023-ETO}. As we illustrate in Example~\ref{exam1}, the intuitive combination of an accuracy-oriented predictive model and a downstream policy can yield counterintuitive and suboptimal outcomes. 

\subsubsection{Limitations of ETO: a motivating example}
\begin{figure}[h]
    \includegraphics[width=0.98\textwidth]{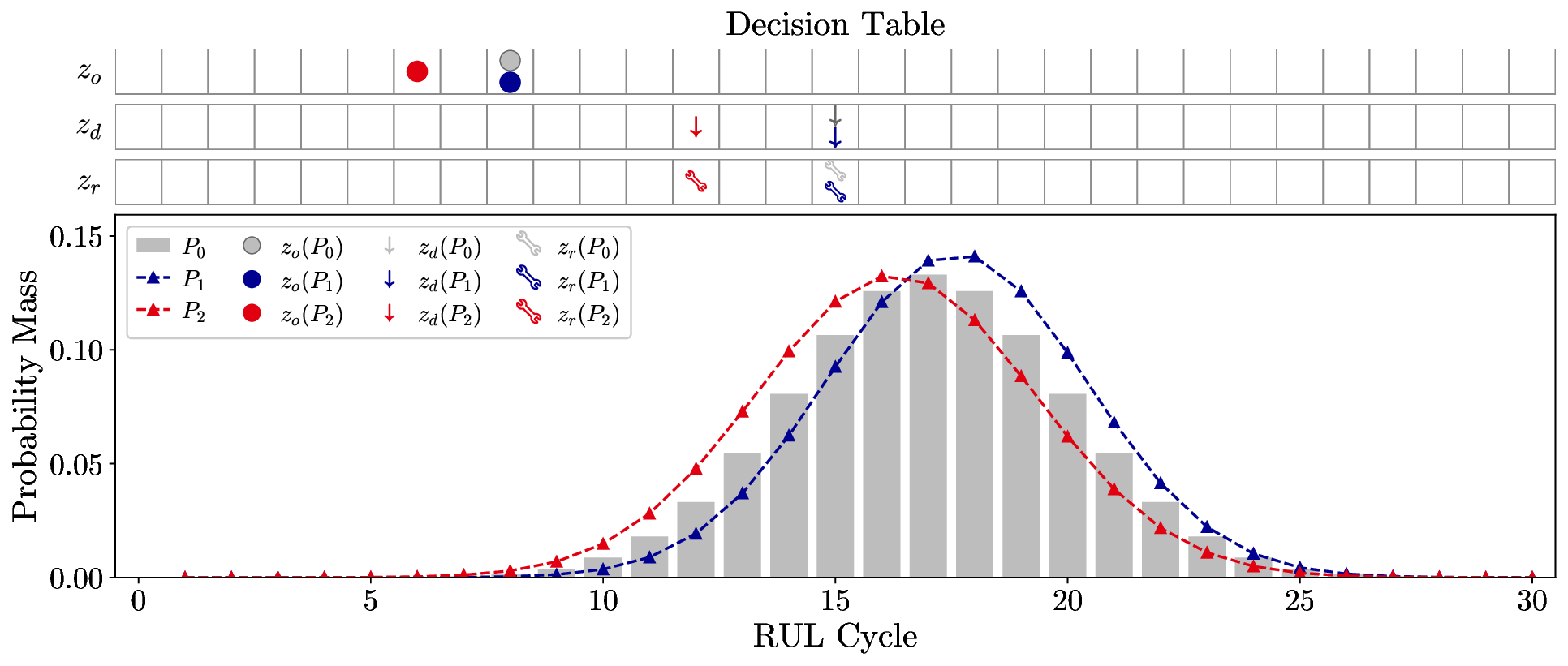}
    \caption{
    Target RLD $P_0$, two estimates $P_1$ and $P_2$ with identical KL-div from $P_0$, and their corresponding ordering, decommissioning, and replacement decisions in Example~\ref{exam1}. Although $P_1$ and $P_2$ have identical estimation accuracy, they induce different maintenance decisions.}\label{fig1}
\end{figure}

To illustrate how the ETO approach can be suboptimal in maintenance, we construct a small example in which two RLD estimates $P_1$ and $P_2$, having identical accuracy in the sense of KL-div compared to a target (true) distribution $P_0$, can lead to different decisions $\boldsymbol{z}_1$ and $\boldsymbol{z}_2$, respectively. Furthermore, the two decisions lead to distinct expected costs under the true RLD $P_0$.
\begin{example}\label{exam1}
    Consider an instance $H = 30,\,c_h = 1,\, c_p = 5,\, c_f = 5 c_p,\,\boldsymbol{\tau} \sim \mathrm{Uniform}\lrc{3,4,5,6,7}$, and the feasible set $\mathcal{Z}_o = \mathcal{Z}_d = \{0,1,2,\dots,30\}, \mathcal{Z}_r = \{3,6,\dots,30\}$. We assume two estimates of the target RLD $P_0$, denoted by $P_1$ and $P_2$, are generated by two distinct prognostic algorithms. Specifically, the accuracies of the two estimates are imperfect yet equivalent, i.e., $\mathbb{D}_{\mathrm{KL}}(P_0 \Vert P_1) = \mathbb{D}_{\mathrm{KL}}(P_0\Vert P_2) = \delta > 0$. The probability masses of these distributions are illustrated in Figure~\ref{fig1}. Denote the corresponding decision $\boldsymbol{z}_i = g(P_i),\ i = 0,1,2$. We compute the regret of decisions $\boldsymbol{z}_1$ and $\boldsymbol{z}_2$ relative to $\boldsymbol{z}_0$ as the metric of decision quality, which is defined as follows:
    \begin{equation*}\nonumber
        \text{Reg}(P_i, P_0) = \mathbb{E}_{P_0}\left[c(g(P_i) ; \bxi) - c(\boldsymbol{z}_0;\bxi)\right], \quad i = 1,2.
    \end{equation*}
    Figure~\ref{fig1} also portrays the scheduling of decisions under the true distribution $P_0$, estimates $P_1$ and $P_2$. Clearly, the regret of $P_1$ is 0 because $\boldsymbol{z}_1 = \boldsymbol{z}_0$ whereas that of $P_2$ is nonzero. Therefore, two estimates with identical estimation accuracy, in the sense of KL-div, lead to decisions that have distinct outcomes.
\end{example}

Example~\ref{exam1} illustrates that identical estimation accuracy under the KL-div criterion can nevertheless lead to different decision quality when the downstream optimization model uses the estimate as input. Moreover, an imperfect estimate does not necessarily result in a suboptimal decision, as demonstrated by the fact that $P_1$ yields the same decision $\boldsymbol{z}_1 = \boldsymbol{z}_0$. In the following section, we generalize this phenomenon to a broader class of problems.

\subsubsection{Lipschitz condition for ETO}
In many practical implementations of the ETO approach in data-driven PdM, the following Lipschitz continuity property in Definition~\ref{def:1} is implicitly assumed, which motivates the effort to improve the estimation accuracy as much as possible.
\begin{definition}[Lipschitz Continuity]\label{def:1}
    A decision mapping $g$ is Lipschitz continuous on $\mathcal{P}_{\Xi}$ with respect to the total variation distance $\Vert \cdot \Vert_{\operatorname{TV}}$, if $\exists L_{kl} > 0,\forall P_1, P_2 \in \mathcal{P}_{\Xi}$,
    \begin{equation*}
        \Vert g(P_1) - g(P_2) \Vert_1 \leq L_{kl} \left\Vert P_1 - P_2 \right\Vert_{\operatorname{TV}} .
    \end{equation*}
    Additionally, the right-hand side term can be further bounded by Pinsker's inequality~\cite[Lemma 2.5]{Pinsker}:
    \begin{equation*}
         \left\Vert P_1 - P_2 \right\Vert_{\operatorname{TV}} 
         \coloneqq \frac{1}{2}\sum_{\xi \in \Xi} |P_1(\xi) - P_2(\xi)|  
         \leq \min\left\{
            \sqrt{\frac{1}{2}\mathbb{D}_{\mathrm{KL}}(P_1 \Vert P_2)}, \sqrt{\frac{1}{2}\mathbb{D}_{\mathrm{KL}}(P_2 \Vert P_1)}
        \right\}.
    \end{equation*}
\end{definition}

If $g$ is Lipschitz continuous on $(\mathcal{P}_\Xi, \Vert \cdot \Vert_{\operatorname{TV}})$, it endorses the ETO approach as follows: Consider $P_1$ as some conditional $\mathbb{P}_{\bxi\mid X}$ and $P_2$ its estimate which is generated under the KL-div criterion. As $\mathbb{D}_{\mathrm{KL}}(P_1 \Vert P_2)$ decreases, Pinsker's inequality tightens the upper bound on $\left\Vert P_1 - P_2 \right\Vert_{\operatorname{TV}} $; under Lipschitz continuity, this would in turn control the deviation in the induced decision. However, for a broad range of decision-making problems that satisfy Assumptions \ref{A1}--\ref{A2}, we show in Proposition~\ref{prop1} that the Lipschitz continuity does not hold.

\begin{assumption}\label{A1}
    The number of support points $|\Xi| \geq 2$, and the number of feasible decisions $|\mathcal{Z}|\geq 2$.
\end{assumption}
\begin{assumption}\label{A2}
    For each feasible maintenance decision $\bz$, let $C_{\bz}$ denote its scenario-cost vector:
    \begin{equation*}
        C_{\bz} \coloneqq \lra{ c(\bz;\xi^1),\dots, c(\bz;\xi^H) }.
    \end{equation*}
    Let $S_{\bz}$ denote the optimal decision region of RLDs under which $\bz$ is optimal:
    \begin{equation*}
        S_{\bz} \coloneqq \left\{P \in \mathcal{P}_{\Xi}: \mathbb{E}_P[c(\boldsymbol{z};\bxi)] \leq \mathbb{E}_P[c(\boldsymbol{z}';\bxi)],\forall \boldsymbol{z}' \in \mathcal{Z}  \right\}.
    \end{equation*}
    There exist two feasible decisions, denoted $\bz_{\star}$ and $\bz_\dagger$, satisfying both conditions: (a) The scenario-cost vectors are different, i.e., $\Vert C_{\bz_\star} - C_{\bz_\dagger} \Vert \neq 0$; (b) Each decision is optimal under at least one strictly positive RLD. That is, both $S_{\bz_\star}$ and $S_{\bz_\dagger}$ intersect the relative interior of the probability simplex (probability distributions that assign positive probability to every support point).
\end{assumption}

Here, Assumptions~\ref{A1}--\ref{A2} are imposed to avoid trivial cases. Specifically, Assumption~\ref{A1} ensures that neither the RUL support nor the feasible decision set is a singleton. Assumption~\ref{A2} excludes an impractical case where for every distribution $P$ in the relative interior of $\mathcal{P}_\Xi$, the optimal decision is always the same. Such a case may arise, for instance, when a sufficiently large downtime penalty is considered, so the ordering and replacement decisions should be scheduled as soon as possible. 

With the two assumptions, it suffices to show that $g$ is not continuous with respect to $P$ to conclude that the Lipschitz continuity of $g$ over $(\mathcal{P}_\Xi, \Vert \cdot \Vert_{\operatorname{TV}})$ does not hold. We formally present it in Proposition~\ref{prop1}.

\begin{proposition}\label{prop1}
    Suppose the optimization model $g$ satisfies Assumptions~\ref{A1}--\ref{A2}. Then, $\forall L_{kl} > 0$, there exist $P_1, P_2 \in \mathcal{P}_{\Xi}, \,P_1 \neq P_2$, such that 
    \begin{equation*}
        \Vert g(P_1) - g(P_2) \Vert_1 
        > L_{kl} \min \left\{
        \sqrt{\frac{1}{2}\mathbb{D}_{\mathrm{KL}}(P_1 \Vert P_2)}, \sqrt{\frac{1}{2}\mathbb{D}_{\mathrm{KL}}( P_2 \Vert P_1)} 
        \right\}
        \geq L_{kl} \,\Vert P_1 - P_2 \Vert_{\operatorname{TV}}.
    \end{equation*}
    Therefore, $g$ is not Lipschitz continuous over $(\mathcal{P}_\Xi, \Vert \cdot \Vert_{\operatorname{TV}})$.
\end{proposition}

\begin{figure}
    \centering
    \includegraphics[width=0.95\linewidth]{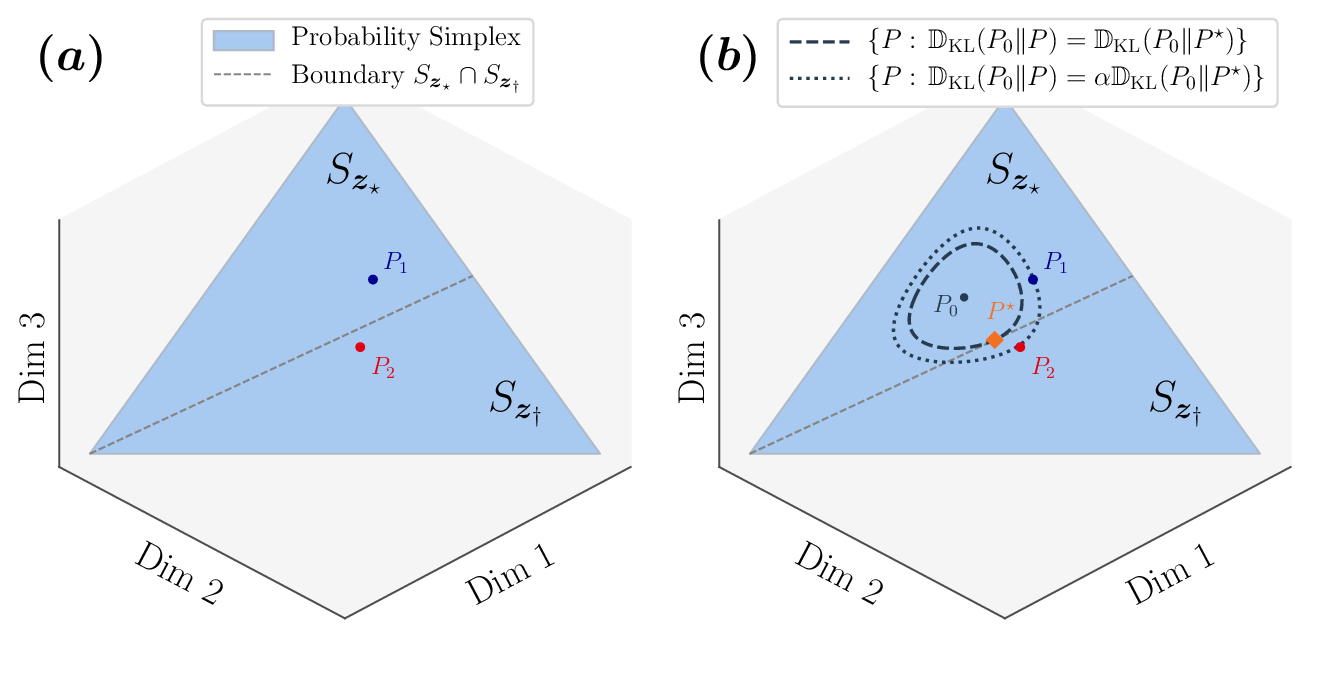}
    \caption{
    Geometric illustration of the mismatch between RLD estimation accuracy and maintenance decisions. (a) Two close RLD estimates $P_1$ and $P_2$ can lie in adjacent decision regions and induce different optimal decisions, as characterized in Proposition~\ref{prop1}. (b) Two RLD estimates can have identical KL-div from the true RLD $P_0$ while lying in different decision regions and therefore inducing different maintenance decisions, as discussed in Remark~\ref{remark1}.
    }
    \label{propfig}
\end{figure}

\begin{proofof}{Proposition}{prop1}
        The proof can be found in~\ref{proofofProposition1}. To give a sketch of the proof, consider that $P_1$ and $P_2$ reside in two adjacent optimal decision regions, such that $P_1 \in S_{\bz_\star}, P_2 \in S_{\bz_\dagger}$ and $\bz_\star \neq \bz_\dagger$. Furthermore, as $\mathbb{D}_{\mathrm{KL}}(P_1 \Vert P_2) = \delta$ converges to 0, $P_1$ and $P_2$ converge to the boundary between $S_{\bz_\star}$ and $S_{\bz_\dagger}$. However, for any $\delta > 0$, $P_1$ and $P_2$ are still within their respective regions, resulting in distinct decisions $g(P_1) \neq g(P_2)$. A geometric illustration of Proposition~\ref{prop1} is given in Figure~\ref{propfig}(a) where the boundary between $S_{\bz_\star}$ and $S_{\bz_\dagger}$ is assumed to be $S_{\bz_\star} \bigcap S_{\bz_\dagger} := \{P \in \mathcal{P}_\Xi \mid 2 P(\xi^1) +3P(\xi^2) + P(\xi^3) = 2 \}$.
\end{proofof}


\begin{remark}\label{remark1}
Proposition~\ref{prop1} reveals that the global Lipschitz continuity of $g$ with respect to the total variation distance does not hold when $g$ satisfies Assumptions~\ref{A1}--\ref{A2}. The core reason is that the discrete feasible set $\mathcal{Z}$ partitions $\mathcal{P}_{\Xi}$ into multiple adjacent polytopes $S_{\bz} \subset \mathcal{P}_{\Xi}$, within each of which a particular decision remains optimal. We further illustrate this geometric insight in Figure~\ref{propfig}(b), where we assume that there are only two optimal decision regions, $S_{\bz_\star}$ and $S_{\bz_\dagger}$, and that the true RLD $P_0$ resides in $S_{\bz_\star}$. In Figure~\ref{propfig}(b), we define
$
    P^\star = \argmin_{P \in S_{\bz_\star} \cap S_{\bz_\dagger}}
    \mathbb{D}_{\mathrm{KL}}(P_0 \Vert P),
$
which is the KL-divergence projection of $P_0$ onto the common boundary $S_{\bz_\star} \cap S_{\bz_\dagger}$. Consider two estimates of $P_0$, denoted by $P_1$ and $P_2$. If both estimates satisfy
    $
    \mathbb{D}_{\mathrm{KL}}(P_0 \Vert P_i)
    < \mathbb{D}_{\mathrm{KL}}(P_0 \Vert P^\star),
    \, i=1,2,$
then both must remain in $S_{\bz_\star}$ and therefore incur zero decision regret. However, once the estimation-error level exceeds $\mathbb{D}_{\mathrm{KL}}(P_0 \Vert P^\star)$, e.g.,
$\alpha\,\mathbb{D}_{\mathrm{KL}}(P_0 \Vert P^\star)$ for some $\alpha>1$, there exist estimates at this error level that reside in $S_{\bz_\dagger}$. Consequently, estimation accuracy alone no longer determines decision quality; the geometry of the estimate relative to the decision-region boundary also becomes critical.
\end{remark}

In a data-driven setting where a parametric prognostic model is adopted, an estimation method cannot generally guarantee a sufficiently small error for almost every conditional RLD due to limited data and model misspecification. Consequently, as \cite{Kamariotis-2024-RESS} demonstrated empirically, the training hyperparameters have a non-negligible effect on the downstream decision quality in the ETO approach. Next, we formally clarify the mismatch between RLD estimation and maintenance optimization based on the concept of consistency and introduce an approach that guarantees the consistency of the learning process.

\subsection{Definition of learning consistency}
To understand the link between RLD estimation accuracy and maintenance consequences, we contrast two notions of risk. First, \textit{decision risk} $R(\omega)$, also referred to as true risk, is the expected maintenance cost induced by model parameter $\omega$ when applying the corresponding policy $\pi_\omega$. We define the decision risk $R(\omega)$ and its empirical counterpart $\hat{R}(\omega)$ as:
\begin{align*}
    R(\omega) &\coloneqq \mathbb{E}_{\P}\lrb{c( \pi_\omega(X); \boldsymbol{\xi} )} \\
    \hat{R}(\omega) &\coloneqq \frac{1}{n} \sum_{i=1}^{n} c(\pi_\omega(x_i);\xi_i).
\end{align*}
Second, \textit{estimation risk} $R_\ell(\omega)$, also known as surrogate risk, measures how well the model estimates under a user-defined criterion $\ell$ independently of any decision-making problem:
\begin{equation}
    R_\ell(\omega) \coloneqq \mathbb{E}_{\P} \lrb{\ell(m(X;\omega),\bxi)}.\label{estimation-risk}
\end{equation}
Specifically, the objective function in Equation~\eqref{functional-ETO} (with $\lambda = 0$) corresponds to the empirical counterpart of estimation risk in Equation~\eqref{estimation-risk}. Under some standard assumptions, the empirical estimation risk minimizer $\hat{\omega}_{eto}$, depending on the model complexity and the size of the training set, admits a generalization guarantee for the estimation risk~\citep{Bartlett-2021-Deeplearning}. However, the guarantee on estimation risk cannot directly translate into guarantees on decision risk. In the PdM context, a lower empirical loss on $\mathcal{D}$ does not guarantee lower maintenance cost when $m$ and $g$ are deployed jointly. As discussed in Section~\ref{sec:ETO}, this misalignment arises when the decision map $g$ lacks the Lipschitz continuity in RLD.

We proceed to formally define the concept of model misspecification.

\begin{definition}[model misspecification]\label{def:mis}
    A model class $\mathcal{M}$ is misspecified with respect to $\P$ if
    \begin{equation*}
    \forall \, \omega \in \Omega, \quad \int_\mathcal{X} \mathbb{D}_{\mathrm{KL}} \lra{\P_{\bxi \mid X=x} \Vert m(x;\omega)} \text{d}\P_{X}(x) > 0,\notag
    \end{equation*}
    i.e., no model in $\mathcal{M}$ can achieve perfect estimation for the true distribution $\mathbb{P}$. By contrast, a model class $\mathcal{M}$ is well-specified with respect to $\P$ if
    \begin{equation*}
    \exists \, \omega \in \Omega, \quad \int_\mathcal{X} \mathbb{D}_{\mathrm{KL}} \lra{\P_{\bxi \mid X=x} \Vert m(x;\omega)} \text{d}\P_{X}(x) = 0.\notag
    \end{equation*}
\end{definition}

Model misspecification is an unavoidable aspect of data-driven methodologies. Rather than assuming access to a well-specified model class, it is more critical to understand how misspecification influences decision-making when the learning objective is to identify a best-in-class, accuracy-oriented estimator under limited data. To rigorously characterize this relationship, we adopt the notion of \textit{consistency} introduced by \cite{consistency}. This notion formalizes the extent to which improvements in predictive performance translate into better decision quality.
\begin{definition}[$\P$-consistency]
    A loss function $\ell$ is $\P$-consistent if for all $\epsilon > 0$, there exists a $\gamma > 0$ (which may depend on $\epsilon$), such that: 
\begin{equation*}
    \begin{aligned}
        \set{\omega \mid R_\ell(\omega) - \inf_{\omega' \in \Omega} R_\ell(\omega') \leq \gamma } \subseteq 
        \set{\omega \mid R(\omega) - \inf_{\omega' \in \Omega} R(\omega') \leq \epsilon }.
    \end{aligned}
\end{equation*}
\end{definition}
The $\mathbb{P}$-consistency condition describes how a loss $\ell$ supports decision making for a specified model class $\mathcal{M}$, optimization model $g$, and data distribution $\mathbb{P}$. If $\ell$ is $\mathbb{P}$-consistent, then minimizing the risk induced by $\ell$ yields a decision risk no worse than that of the best decision-oriented model in the same model class, up to the prescribed tolerance $\epsilon$. Nonetheless, establishing $\mathbb{P}$-consistency for general losses is challenging as it hinges on the specific structure of the downstream optimization problem.

Despite the difficulty of analyzing the consistency of a surrogate loss, there exists at least one $\P$-consistent loss function $L$ defined in Equation~\eqref{decision-loss}, which replaces the estimation error by the decision loss evaluated by plugging the policy output into the underlying cost function for any sample $(x,\xi)$:
\begin{equation}
    L(m(X;\omega), \xi) \coloneqq c(g(m(X;\omega));\xi) = c(\pi_\omega(X);\xi). \label{decision-loss}
\end{equation}
We show that this specific loss $L$ is $\P$-consistent by the following Observation~\ref{L-consistent}.

\begin{observation}\label{L-consistent}
    The loss function $L$ in Equation~\eqref{decision-loss} is $\P$-consistent for any model class $\mathcal{M}$, since by definition $R_L(\omega) = R(\omega)$. Indeed, for all $\epsilon > 0$, one can take any $ 0< \gamma \leq \epsilon$, such that the following statement always holds:
\begin{equation*}
    \set{\omega \mid R_L(\omega) - \inf_{\omega' \in \Omega} R_L(\omega') \leq \gamma } 
    \subseteq 
    \set{\omega \mid R(\omega) - \inf_{\omega' \in \Omega} R(\omega') \leq \epsilon}.
\end{equation*}
\end{observation}

\subsection{Non-asymptotic guarantee of consistency}\label{sec:3_3}
A critical challenge remains in minimizing the risk $R_L(\omega)$. Observation~\ref{L-consistent} suggests that minimizing the risk induced by $L$ is equivalent to minimizing the true risk, but it does not suggest any practical approach. Since the true distribution \( \mathbb{P} \) is unknown, models have to be learned using a finite dataset \( \mathcal{D} \). 
The key concern is then practical: how can a parameter $\omega$ learned from finite data $\mathcal{D}$ approximate a minimizer of $R_{L}(\omega)$? To address this gap, we leverage the framework of Probably Approximately Correct (PAC) learning from computational learning theory to derive probabilistic guarantees for a model parameter $\omega$ that minimizes the empirical decision risk over $\mathcal{D}$. We leverage the following assumptions to derive a non-asymptotic guarantee for the learning process.
\begin{assumption}\label{A3}
     The dataset $\mathcal{D}$, containing sensor data and RUL pairs, is generated i.i.d.~from a data-generating process $\P$. 
     Denote the empirical measure by $\P^n \coloneqq \frac{1}{n}\sum_{i\in[n]}\delta_{(X_i,\bxi_i)}$ with $\delta_{(\cdot,\cdot)}$ the Dirac measure, and $(X_i,\bxi_i) \sim \P$ for all $i = 1,\dots,n$.
\end{assumption}
\begin{assumption}\label{A4}
    The maintenance decision feasible set $\mathcal{Z}$ is finite, i.e., $|\mathcal{Z}| = K < \infty$.
\end{assumption}
\begin{assumption}\label{A5}
     The maintenance cost function $c$ in Equation~\eqref{obj} is bounded, i.e., for every maintenance decision vector $\boldsymbol{z} \in \mathcal{Z}$ and for all RUL scenarios $\xi \in \Xi$, $c(\boldsymbol{z};\xi) \in [C_{l}, C_{u}]$.
\end{assumption}

A model complexity measure is required within the PAC-learning framework. In this work, we leverage the \textit{Natarajan dimension}, which is a multiclass generalization of Vapnik–Chervonenkis dimension. The Natarajan dimension is an appropriate tool to analyze the complexity of policy class $\Pi$, which is a family of mappings from contextual observations $x\in\mathcal X$ to feasible decisions $\bz\in\mathcal Z$. The formal definition of Natarajan dimension is provided in~\ref{sec:B_2}.

\begin{assumption}\label{A6}
    The Natarajan dimension $d_\Pi$ of policy class $\Pi$, which is determined by the model class $\mathcal{M}$ and optimization model $g$, is finite, i.e., there exists some $D_\Pi$ such that $d_\Pi \leq  D_\Pi < \infty$.
\end{assumption}
The finite Natarajan dimension in Assumption~\ref{A6} controls the complexity of the induced prognosis-then-maintenance policy class. Although the feasible decision set $\mathcal{Z}$ is finite, an unrestricted policy class may still realize arbitrarily rich context-dependent decision patterns and therefore need not generalize from finite data. The condition $d_{\Pi} \leq D_{\Pi} < \infty$ restricts this combinatorial complexity and enables uniform convergence between the empirical and population decision risks. In Section~\ref{sec:4_1}, we verify Assumption~\ref{A6} for the policy class $\Pi$ induced by our specified prognostic model class $\mathcal{M}$ and optimization model $g$.

With these assumptions, we establish a non-asymptotic guarantee for our framework in Theorem \ref{theorem-1}. Conceptually, our analysis views the prognosis-then-maintenance procedure as a discrete choice problem over a finite feasible set, where the utility associated with each decision vector is modeled as an affine function in the predicted probability vector. Beyond the joint ordering-replacement problem, our theoretical findings extend to other maintenance problems that share the structural properties specified in our assumptions. 

\begin{theorem}\label{theorem-1}
    Suppose Assumptions \ref{A3}--\ref{A6} hold. Let $C = C_u - C_l$, $K= |\mathcal{Z}|$, and let $D_\Pi < \infty$ be an upper bound on the Natarajan dimension of $\Pi$.
    Then, for any $\delta \in (0,1)$, with probability at least $1 - \delta$ over the draw of $\mathcal{D}$ from $\P^n$, the following inequality holds simultaneously for all $\omega \in \Omega$:
    \begin{equation*}
    R(\omega) \leq 
    \hat{R}(\omega) + C \sqrt{
    \frac{8}{n} \lra{
    D_\Pi \log(2n) + 2 D_\Pi \log K + \log \frac{2}{\delta}
    } }.
    \end{equation*}
\end{theorem}
\begin{proofof}{Theorem}{theorem-1}
    The proof can be found in~\ref{proofoftheorem1}. \qed
\end{proofof}

Theorem \ref{theorem-1} states the high-probability guarantee on the true risk $R(\omega)$ of any model parameter $\omega$, given its empirical risk $\hat{R}(\omega)$, regardless of how $\omega$ is determined. The following Corollary~\ref{corollary-1} further provides the consistency of the learning process, showing that a model parameter learned by minimizing $\hat{R}(\omega)$ approaches the best-in-class decision-risk minimizer with high probability. Thus, the consistency of the learning process is guaranteed.

\begin{corollary}\label{corollary-1}
Let $\hat{\omega}_L$ be one minimizer of $\hat{R}(\omega)$ and $\omega^\star$ be one minimizer of $R(\omega)$.
Suppose Assumptions~\ref{A3}--\ref{A6} hold. Then, for any $\delta \in (0,1)$, with probability at least $1-\delta$ over the draw of $\mathcal{D}$ from $\P^n$,
\begin{equation*}
R(\hat{\omega}_L) - R(\omega^\star) \leq 2C \sqrt{
\frac{8}{n} \lra{ D_\Pi \log(2n) + 2D_\Pi \log K + \log\frac{4}{\delta} }
},
\end{equation*}
where $C = C_u - C_l$. Consequently, for fixed $D_\Pi$ and $K$, the excess risk $R(\hat{\omega}_L) - R(\omega^\star)$ converges to zero at the rate $\mathcal{O}\lra{\sqrt{\log n/n}}$.
\end{corollary}
\begin{proofof}{Corollary}{corollary-1}
    The proof can be found in~\ref{profofcoro1}. \qed
\end{proofof}
Corollary~\ref{corollary-1} establishes consistency of the empirical decision-risk minimizer. As $n\to\infty$, its excess risk converges to zero at rate $O(\sqrt{\log n/n})$, implying convergence to the best-in-class true-risk minimizer.

\subsection{Bridging theory and practice}
We leverage the i.i.d.~condition in Assumption~\ref{A3} to establish Theorem~\ref{theorem-1}. However, in maintenance practice, a contextual observation $x$ may not be a stand-alone observation but instead a time-series sequence constructed using a sliding-window approach~\citep{Li-2018-RESS, Nguyen-2019-RESS}. To bridge this gap and broaden the applicability of our framework, we introduce Assumption~\ref{A7} and derive a convergence result for sliding-window-dependent data in Theorem~\ref{theorem-2}.
\begin{assumption}[Sliding-window-dependent sampling]\label{A7}
    Let $W \in \mathbb{N}$ be a fixed window length and let $\lrc{\lra{U_t, \bxi_t}}_{t \in \mathbb{Z}}$ be an i.i.d.~sequence drawn from some distribution $\mathbb{Q}$. The contextual observation at time $t$ is constructed by a stride-one sliding window: $X_t = (U_{t-W+1}, \dots, U_t)$ where $U_t$ denotes the random variable of sensor observations. The resulting sequence $\lrc{\lra{X_t, \bxi_t}}_{t=1}^{n}$ has a common marginal distribution $\P$ and is identically distributed and $(W-1)$-dependent; that is, for any two index sets \(I,J\subseteq\{1,\ldots,n\}\) satisfying $\min_{ i \in I, j \in J} |i-j| \geq W$, the collections $\lrc{\lra{ X_i, \bxi_i }:i \in I}$ and $\lrc{\lra{ X_j, \bxi_j }:j \in J}$ are independent.
\end{assumption}
Assumption~\ref{A7} captures the dependence induced by overlapping sliding windows: neighboring samples may share up to $W-1$ sensor observations, whereas samples whose indices differ by at least $W$ contain no overlapping sensor observations and are independent under Assumption~\ref{A7}. Theorem~\ref{theorem-2} shows that the i.i.d.~condition in Assumption~\ref{A3} can be relaxed to this sliding-window dependence while preserving a convergence rate of $\mathcal{O}(\sqrt{\log n/n})$ for fixed $W$.

\begin{theorem}\label{theorem-2}
Suppose Assumptions~\ref{A4}--\ref{A7} hold and $n \geq W$. Then, for any $\delta \in (0,1)$, with probability at least $1-\delta$ over the draw of $\mathcal{D}$ induced by the sliding-window sampling process in Assumption~\ref{A7}, the following inequality holds simultaneously for all $\omega \in \Omega$:
\begin{equation*}
     R(\omega) - \hat{R}(\omega) \leq 
     C \sqrt{ \frac{8W}{n} \lra{ D_\Pi \log \lra{2\left\lceil \frac{n}{W}\right\rceil} + 2 D_\Pi \log K + \log\frac{2W}{\delta} } },
\end{equation*}
where $C = C_u-C_l$. Consequently, when $W$, $D_\Pi$, and $K$ are fixed, the generalization gap converges to zero at the rate $\mathcal{O}\lra{\sqrt{\log n / n}}$.
\end{theorem}
\begin{proofof}{Theorem}{theorem-2}
The proof can be found in~\ref{proofoftheorem2}.\qed
\end{proofof}
Theorem~\ref{theorem-2} extends the uniform out-of-sample guarantee in Theorem~\ref{theorem-1} to training samples constructed from overlapping sliding windows under Assumption~\ref{A7}. The dependence among overlapping windows yields an effective amount of independent information of order $n/W$, resulting in a looser finite-sample bound than in the i.i.d.~case. Nevertheless, for fixed $W$, $D_{\Pi}$, and $K$, the convergence rate remains
$\mathcal{O}(\sqrt{\log n/n})$, and the bound reduces to that of Theorem~\ref{theorem-1} when $W=1$. Therefore, the dependence induced by overlapping sliding windows can be accommodated without changing the asymptotic convergence rate of the decision-risk guarantee.

Overall, our theoretical results provide finite-sample guarantees for the decision risk and establish consistency as the sample size $n$ increases. However, the amount of available data in data-driven PdM is typically limited, which motivates evaluating the IEO framework in practical data-driven PdM settings rather than relying solely on asymptotic behavior. Having established the consistency of the loss function $L$, we now turn to the empirical minimization of this loss. In Section~\ref{sec:4}, we develop a practical learning algorithm for implementing the IEO framework.
\section{A Practical IEO Framework}\label{sec:4}
Section~\ref{sec:3} established that minimizing the empirical decision risk is consistent with minimizing the true risk, but left open how this minimization can be carried out computationally. This section addresses that challenge by instantiating the IEO framework with a specific prognostic model and the joint ordering-replacement model. In Section~\ref{sec:4_1}, we propose an LSTM model with MC dropout to obtain flexible RLD estimates. In Section~\ref{sec:4_2}, we develop a central-difference gradient estimator in the one-dimensional RUL space, avoiding direct differentiation through the discontinuous decision map.

\subsection{Prognostic model}\label{sec:4_1}
The prognostic model $m(\cdot;\omega)$ should produce, for each sensor observation $x$, an estimated probability vector $\hat{P}\in\mathcal{P}_{\Xi}$ over the discrete support $\Xi$. The dimension of this output equals the horizon length $H$, which can be large in practice ($H=150$ in our experiments). Various probabilistic prognostic methods have been developed in data-driven PdM to estimate RLDs from sensor observations. In this work, we adopt MC dropout~\citep{Dropout} to construct these estimates. Beyond its regularization effect during training, MC dropout offers two advantages in our setting. First, it preserves a scalar point-prediction output, avoiding an $H$-dimensional output layer with high complexity. Second, unlike approaches that impose a prespecified parametric family on the RLD, such as Gaussian or Weibull distributions, MC dropout provides a flexible distributional approximation and therefore reduces the risk of distributional misspecification discussed in Definition~\ref{def:mis}.

To maintain consistency with our theoretical analysis in Section~\ref{sec:3}, we interpret $m(x;\omega)$ as the dropout-induced RLD obtained by marginalizing over the dropout randomness. Let $\mathcal{B}\coloneqq\{0,1\}^{d_b}$ denote the finite dropout-mask space, where $d_b$ is determined by the model architecture, and let $p_r$ denote the probability mass function on $\mathcal{B}$ induced by dropout rate $r$. We write $\mathbf{b}\sim p_r$ for a sampled dropout mask. The dropout-induced RLD therefore is
\begin{equation*}
     m(x;\omega)
    \coloneqq
    \left[
        \mathbb{P}_{\mathbf{b}\sim p_r}
        \lra{m(x;\omega,\mathbf{b})\in I_1},
        \ldots,
        \mathbb{P}_{\mathbf{b}\sim p_r}
        \lra{m(x;\omega,\mathbf{b})\in I_H}
    \right]^\top
    \in\mathcal{P}_{\Xi},
\end{equation*} 
where $m(x;\omega,\mathbf{b})$ denotes the scalar RUL prediction under dropout mask $\mathbf{b}$. Thus, $m(x;\omega)$ is the exact model-implied RLD, corresponding to the limiting distribution obtained from infinitely many stochastic forward passes. In implementation, however, $m(x;\omega)$ is approximated using $S$ stochastic forward passes. Given independently sampled masks $\mathbf{b}_1,\ldots,\mathbf{b}_S\overset{\mathrm{i.i.d.}}{\sim}p_r$, we construct the empirical RLD
\begin{equation*}
\begin{aligned}
    m(x;\omega)& = \left[ \hat{P}(\xi^1),\ldots,\hat{P}(\xi^H) \right]^\top,\\
    \hat{P}(\xi^h) &\approx\frac{1}{S} \sum_{s=1}^{S}
    \mathbf{1}\left[ m(x;\omega,\mathbf{b}_s) \in I_h \right], \quad h=1,\ldots,H.
\end{aligned}
\end{equation*}
We take $S=1000$ MC passes in this work to mitigate the sampling variability and found this choice produces stable RLD estimates in our numerical experiments.

The choice of prognostic architecture is not central to our framework. Therefore, we adopt an LSTM because it is well suited to processing sequential observations and provides a competitive baseline for RUL prognostics. In particular, the sliding-window input structure used by the LSTM is consistent with the dependent-data setting considered in Theorem~\ref{theorem-2}. For theoretical completeness, we further show that the policy class $\Pi$ induced by the LSTM-dropout model class and optimization model $g$ has finite Natarajan dimension, thereby verifying Assumption~\ref{A6} for this specification. For brevity, the formal statement is deferred to Proposition~\ref{prop:3} in~\ref{propLSTM}.
\subsection{Central-difference gradient descent}\label{sec:4_2}
This section develops a tractable learning procedure for the IEO framework. Rather than training the prognostic model to improve estimation accuracy, we train it to improve the quality of downstream maintenance decisions. The main difficulty is that both the construction of the dropout-induced RLD and the downstream optimization model involve discrete operations, which prevents direct gradient-based training under the decision loss. We therefore develop a surrogate training procedure in the one-dimensional RUL space using central differences. The resulting algorithm is summarized in Algorithm~\ref{alg:1}. For simplicity, Algorithm~\ref{alg:1} presents a per-sample formulation, whereas the experiments use a mini-batch version.

By replacing the loss function $\ell_{eto}$ in Equation~\eqref{functional-ETO} by the decision loss $L$ in Equation~\eqref{decision-loss}, the IEO framework targets the following empirical decision-risk minimization problem:
\begin{equation}
    \hat{\omega}_{ieo}
    \coloneqq
    \argmin_{\omega \in \Omega}
    \hat{R}(\omega)
    =
    \argmin_{\omega \in \Omega}
    \frac{1}{n}
    \sum_{i=1}^{n}
    c\lra{\pi_\omega(x_i);\xi_i}.
    \label{emp:ieo}
\end{equation}
Equation~\eqref{emp:ieo} defines the target learning objective of IEO. However, its gradient cannot be obtained by directly differentiating through the prognosis-then-maintenance policy.

To see this, for each sample $(x_i,\xi_i)$, let
$
    m(x_i;\omega)=\hat{P}_i,
    \,
    g(\hat{P}_i)=\hat{\bz}_i,
    \,
    c(\hat{\bz}_i;\xi_i)=c_i,
$
and define the sample-wise decision loss
$
    L_i(P) \coloneqq c(g(P);\xi_i).
$
If the composed mapping were differentiable, the chain rule would formally give
\begin{equation}
    \nabla_{\omega} c_i
    =
    J_{m,i}(\omega)^\top
    \nabla_{P} L_i(P)
    \big|_{P=\hat{P}_i},
    \label{eq:formal_chain}
\end{equation}
where
\begin{equation*}
    J_{m,i}(\omega)
    \coloneqq
    \frac{\partial m(x_i;\omega)}
    {\partial \omega^\top}
    \in \mathbb{R}^{H\times q}
\end{equation*}
is the Jacobian of the estimated RLD with respect to the model parameters. If the decision map $g$ were also differentiable, the second term on the right-hand side of Equation~\eqref{eq:formal_chain} could be decomposed as
\begin{equation}
    \nabla_{P}L_i(P)
    =
    J_g(P)^\top
    \nabla_{\bz}c(\bz;\xi_i)
    \big|_{\bz=g(P)},
    \label{eq:formal_chain_g}
\end{equation}
where $J_g(P)$ denotes the Jacobian of the decision map with respect to the RLD.

Neither decomposition provides a useful gradient in our setting. First, the exact dropout-induced RLD $m(x_i;\omega)$ in Section~\ref{sec:4_1} can be expressed as a finite weighted sum over $\mathbf b\in\mathcal B$, with weights $p_r(\mathbf b)$, of hard interval-membership indicators. It is therefore piecewise constant with respect to $\omega$. Consequently, $J_{m,i}(\omega)$ is zero almost everywhere and is not defined when, for at least one dropout mask, the scalar prediction crosses an interval boundary. Second, the decision map
$
g:\mathcal{P}_{\Xi}\rightarrow\mathcal{Z}
$
maps a continuous probability vector to the discrete feasible set $\mathcal{Z}$ and is likewise piecewise constant and discontinuous at decision-region boundaries. Hence, $J_g(P)$ is also zero almost everywhere and is not defined at these boundaries. Direct backpropagation through the dropout-induced RLD construction and the optimization model therefore cannot provide a useful gradient signal for training the prognostic model.

To circumvent these discontinuities, we instead operate directly in the one-dimensional RUL space. Conditional on a dropout mask $\mathbf{b}_s$, the scalar stochastic prediction $\hat{\xi}_i = m(x_i;\omega,\mathbf{b}_s)$ remains differentiable with respect to $\omega$ almost everywhere, and its gradient $\nabla_{\omega}m(x_i;\omega,\mathbf{b}_s) $ can be computed using standard automatic-differentiation tools~\citep{autodiff}. Let $\kappa:\mathbb{R}_{+}\rightarrow\{1,\dots,H\}$ denote the RUL-support index mapping, where $\kappa(y)=h$ if $y\in I_h$, with endpoint and upper-tail conventions such that every nonnegative prediction is assigned to exactly one interval. Let $e_h\in\mathcal{P}_{\Xi}$ denote the degenerate RLD assigning probability one to the support point $\xi^h$. For sample $i$, define
\begin{equation*}
    V_i(h)
    \coloneqq
    c(g(e_h);\xi_i),
    \qquad
    h=1,\dots,H,
\end{equation*}
and extend $V_i$ to the scalar RUL space through the piecewise-constant function
\begin{equation*}
    f_i(y)
    \coloneqq
    V_i(\kappa(y))
    =
    c(g(e_{\kappa(y)});\xi_i).
\end{equation*}
The function $f_i$ measures the maintenance cost induced by interpreting the scalar prediction $y$ as a degenerate RLD. Although $f_i$ is discontinuous, we define its one-cycle local average as
\begin{equation*}
    \bar{f}_i(y)
    \coloneqq
    \frac{1}{2}
    \int_{y-1}^{y+1}
    f_i(u)\mathrm{d}u.
\end{equation*}
For an interior RUL interval, the derivative of $\bar{f}_i$ admits the following central-difference representation:
\begin{equation}
    \tilde{g}_i(y)
    \coloneqq
    \frac{1}{2}
    \lrb{
        c(g(e_{\kappa(y)+1});\xi_i)
        -
        c(g(e_{\kappa(y)-1});\xi_i)
    }.
    \label{centraldiff}
\end{equation}
As established later in Proposition~\ref{prop2}, $\tilde{g}_i(y)$ is the exact derivative of the one-cycle locally averaged decision loss for interior RUL intervals. Therefore, instead of differentiating the discrete mappings in Equation~\eqref{emp:ieo}, we use Equation~\eqref{centraldiff} as a surrogate learning signal in the scalar RUL space.

\begin{proposition}\label{prop2}
For any non-integer $y$ satisfying $2\leq\kappa(y)\leq H-1$,
\begin{equation*}
    \frac{\mathrm{d}}{\mathrm{d}y}
    \bar{f}_i(y)
    =
    \frac{1}{2}
    \lrb{
        V_i(\kappa(y)+1)
        -
        V_i(\kappa(y)-1)
    }.
\end{equation*}
Hence, for interior RUL intervals, the central difference in Equation~\eqref{centraldiff} is the exact derivative of the one-cycle locally averaged decision loss.
\end{proposition}

\begin{proofof}{Proposition}{prop2}
For any non-integer $y$ satisfying $2\leq\kappa(y)\leq H-1$, the points $y-1$ and $y+1$ lie inside the domain of $f_i$ and satisfy
$
\kappa(y\pm1)=\kappa(y)\pm1.
$
Since $f_i$ is piecewise constant with possible discontinuities only at integer points, it is integrable and continuous at both endpoints of $[y-1,y+1]$. The fundamental theorem of calculus therefore gives
\begin{align*}
    \frac{\mathrm{d}}{\mathrm{d}y}\bar{f}_i(y)
    &=
    \frac{1}{2}
    \lrb{
        f_i(y+1)-f_i(y-1)
    } \\
    &=
    \frac{1}{2}
    \lrb{
        V_i(\kappa(y)+1)
        -
        V_i(\kappa(y)-1)
    }.
\end{align*}
Substituting
$
V_i(h)=c(g(e_h);\xi_i)
$
completes the proof.
\qed
\end{proofof}

Conditional on a sampled dropout mask $\mathbf{b}$ and for an interior RUL interval, the chain rule gives
\begin{equation}
    \nabla_{\omega}
    \bar{f}_i
    \lra{
        m(x_i;\omega,\mathbf{b})
    }
    =
    \bar{f}'_i
    \lra{
        m(x_i;\omega,\mathbf{b})
    }
    \nabla_{\omega}
    m(x_i;\omega,\mathbf{b}),
    \label{eq:surrogate_chain}
\end{equation}
where the first term is given exactly by Equation~\eqref{centraldiff}, while the second term is obtained through automatic differentiation. Thus, the central-difference procedure provides a tractable stochastic surrogate update for the empirical decision-risk objective in Equation~\eqref{emp:ieo}. At the boundary intervals, we use the corresponding one-sided finite difference. Algorithm~\ref{alg:1} is therefore a surrogate gradient procedure motivated by the empirical decision-risk objective in Equation~\eqref{emp:ieo}.

In implementation, we additionally adopt an ETO warm-start to improve the stability and efficiency of IEO training. Specifically, we initialize the model parameter with the ETO-trained solution $\hat{\omega}_{\mathrm{eto}}$, which provides a predictive model that has already learned the underlying degradation patterns from the data. This initialization avoids starting IEO from an uninformative model and is computationally inexpensive, since ETO training does not require repeatedly solving the downstream optimization problem.

\begin{algorithm}[t]
\caption{\textbf{Central-Difference Surrogate Gradient Descent with Dropout}}
\label{alg:1}
\vskip 0.2cm
\KwIn{Dataset $\mathcal{D}$, prognostic model $m$, learning rate $\eta$, dropout rate $r$, \textbf{CSO} model $g$, ETO-trained parameter $\hat{\omega}_{eto}$.}
\KwOut{Estimated parameter $\hat{\omega}_{ieo}$ obtained by the surrogate procedure for Equation~\eqref{emp:ieo}.}

Initialize model parameter
$
\omega_0 \gets \hat{\omega}_{eto},
\,
k \gets 0.
$

\While{not converged}{
    \For{$i \gets 1$ \KwTo $n$}{
        Draw dropout mask $\mathbf{b}_{i,k}$ with dropout rate $r$.

        Predict
        \[
            \hat{\xi}_{i,k}
            \gets
            m(x_i;\omega_k,\mathbf{b}_{i,k}).
        \]

        Set
        \[
            h_{i,k}
            \gets
            \kappa(\hat{\xi}_{i,k}),
        \]
        and define
        \[
            h_{i,k}^-
            =
            \max\{1,h_{i,k}-1\},
            \qquad
            h_{i,k}^+
            =
            \min\{H,h_{i,k}+1\}.
        \]

        Solve
        \[
            \bz_{i,k}^+
            =
            g(e_{h_{i,k}^+}),
            \qquad
            \bz_{i,k}^-
            =
            g(e_{h_{i,k}^-}).
        \]

        Compute the surrogate gradient
        \[
            \tilde{g}_{i,k}
            \gets
            \frac{
                c(\bz_{i,k}^+;\xi_i)
                -
                c(\bz_{i,k}^-;\xi_i)
            }{
                h_{i,k}^+
                -
                h_{i,k}^-
            }.
        \]

        Update
        \[
            \omega_{k+1}
            \gets
            \omega_k
            -
            \eta
            \tilde{g}_{i,k}
            \nabla_{\omega}
            m(x_i;\omega_k,\mathbf{b}_{i,k}).
        \]

        $k\gets k+1$.
    }
}
\Return $\hat{\omega}_{ieo}\gets\omega_k$\;
\end{algorithm}
\section{Case Studies: Turbofan Engine Maintenance}\label{sec:5}
This section presents a comprehensive computational study of the turbofan engine maintenance problem. We first describe the experimental setup, including the data, model architecture, evaluation metrics, and approaches. We then report two case studies: one under the nominal setting and one under varying levels of risk aversion. All data and code used in this study are publicly available in the accompanying repository\footnote{\texttt{https://github.com/Zhuojun-PhD/DataDrivenMaintenance}}. The experiments were implemented in Python~3.12 using PyTorch~2.11 with CUDA~12.8 and conducted on a single NVIDIA A100 GPU provided by the Mésocentre computing center of Université Paris-Saclay.

\subsection{Experimental setup}
\subsubsection{Data description}
We conduct numerical experiments on the CMAPSS turbofan-engine degradation dataset~\citep{CMAPSS}, which consists of four subsets, \texttt{FD001}--\texttt{FD004}, generated under different operating conditions and fault modes. Each subset provides a training set containing complete run-to-failure trajectories and a test set containing truncated trajectories. Since our maintenance evaluation requires the realized RUL over the complete degradation trajectory, we use only the engines from the original training sets.

For each subset, we consider 100 engines: all 100 engines in \texttt{FD001} and \texttt{FD003}, and the first 100 engines, ordered by the original unit number, in \texttt{FD002} and \texttt{FD004}. We randomly split these engines into training, validation, and test sets 50 times with a ratio of $70/10/20$\footnote{We use random seeds 0-49 in the code.}. The split is performed at the engine level so that observations from the same engine cannot appear in different sets. We further restrict each trajectory to observations with RUL in $[0,99]$ cycles, thereby focusing on the degradation stage in which maintenance decisions are most relevant.

To construct a sample $(x_i,\xi_i)$, we follow the common sliding-window approach for the CMAPSS dataset~\citep{mitici-2023-RESS}. Specifically, a window of length $W=30$ and stride one is used following~\cite{Shang-2022-CIS} to construct the time-series input $x_i$ from the sensor observations preceding the corresponding RUL label $\xi_i$. Among the 21 available sensors, several exhibit negligible variation in \texttt{FD001} and \texttt{FD003}. We therefore remove sensors whose standard deviation is below $10^{-2}$, retaining 14 sensors for \texttt{FD001} and 16 for \texttt{FD003}, while retaining all 21 sensors for \texttt{FD002} and \texttt{FD004}. Each retained sensor is then normalized to $[-1,1]$ according to
\begin{equation*}
    x_{i,j,t} = 2\frac{x_{i,j,t}-\underline{x}_j}{\overline{x}_j-\underline{x}_j} -1, \qquad \forall i,j,t,
\end{equation*}
where $\underline{x}_j$ and $\overline{x}_j$ denote, respectively, the minimum and maximum values of the $j$th sensor extracted from the training set.

\subsubsection{Evaluation metrics}
To evaluate each trained model, we consider two prognostic metrics and two
maintenance metrics over the test set. Let
\[
    \hat{\xi}_{i,s}
    :=
    m(x_i;\omega,\mathbf{b}_{i,s}),
    \qquad s=1,\ldots,S,
\]
denote the $s$th stochastic RUL prediction for sample $(x_i,\xi_i)$ under MC dropout. We use $S=1000$ stochastic forward passes throughout the evaluation. 

For prognostic accuracy, we consider the Mean Absolute Error (MAE) and the Continuous Ranked Probability Score (CRPS). The MAE is computed as
\[
    \mathrm{MAE}(\omega)
    :=
    \frac{1}{nS}
    \sum_{i=1}^{n}\sum_{s=1}^{S}
    \left|\hat{\xi}_{i,s}-\xi_i\right|.
\]
To characterize the direction of probabilistic prediction errors, we decompose the CRPS into underestimation and overestimation components:
\[
    \mathrm{CRPS}^{-}(\omega)
    \coloneqq
    \frac{1}{nS^2}
    \sum_{i=1}^{n}\sum_{s=1}^{S}\sum_{t=1}^{S}
    \left[
        \xi_i-\max\{\hat{\xi}_{i,s},\hat{\xi}_{i,t}\}
    \right]_{+},
\]
and
\[
    \mathrm{CRPS}^{+}(\omega)
    \coloneqq
    \frac{1}{nS^2}
    \sum_{i=1}^{n}\sum_{s=1}^{S}\sum_{t=1}^{S}
    \left[
        \min\{\hat{\xi}_{i,s},\hat{\xi}_{i,t}\}-\xi_i
    \right]_{+}.
\]
Accordingly,
\[
    \mathrm{CRPS}(\omega) = \mathrm{CRPS}^{-}(\omega) + \mathrm{CRPS}^{+}(\omega).
\]
We further define
\[
    \Delta\mathrm{CRPS}(\omega) \coloneqq \mathrm{CRPS}^{-}(\omega) - \mathrm{CRPS}^{+}(\omega),
\]
which measures the directional imbalance of the probabilistic prediction errors. A positive $\Delta\mathrm{CRPS}$ indicates a stronger tendency toward RUL underestimation, whereas a negative value indicates a stronger tendency toward overestimation. Values close to zero indicate a more neutral predictive distribution.

For maintenance performance, we consider Mean Maintenance Cost (MMC) and Mean Failure Rate (MFR). Let $\hat{P}_i$ denote the empirical RLD obtained from the $S$ MC-dropout predictions for $x_i$, and let $\bz_i(\omega) \coloneqq g(\hat{P}_i)$ denote the corresponding maintenance decision. We define
\begin{equation*}
    \begin{aligned}
    \mathrm{MMC}(\omega)
    &\coloneqq
    \frac{1}{n}
    \sum_{i=1}^{n}
    c\bigl(\bz_i(\omega);\xi_i\bigr),\\
    \mathrm{MFR}(\omega)
    &:=
    \frac{1}{n}
    \sum_{i=1}^{n}
    \mathbf{1}\!\left[
    z_{d,i}(\omega)>\xi_i
    \right],
    \end{aligned}
\end{equation*}
    
where $z_{d,i}(\omega)$ denotes the planned decommissioning time prescribed for sample $i$. Thus, MMC measures the average realized maintenance cost, whereas MFR measures the fraction of test samples for which failure occurs before planned decommissioning.

\subsubsection{Models, baselines, and training strategy}
\begin{figure*}[h]
    \centering
    \includegraphics[width=\linewidth]{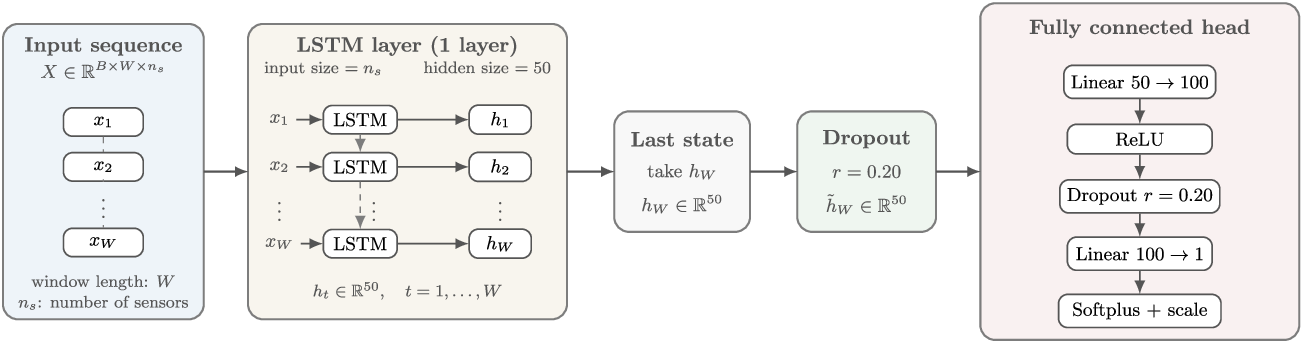}
    \caption{Architecture of the LSTM prognostic model used in the experiments. The final LSTM hidden state is passed through dropout and a fully connected prediction head containing a ReLU activation and a second dropout layer. $B$, $W$, and $n_s$ denote the batch size, sliding-window length, and number of retained sensors, respectively.}
    \label{fig:lstm}
\end{figure*}
For the prognostic task, we use the LSTM architecture illustrated in Figure~\ref{fig:lstm}. The model consists of a single-layer LSTM with hidden dimension 50, followed by a fully connected prediction head. Specifically, the final hidden state is passed through a linear layer from 50 to 100 dimensions, a \texttt{ReLU} activation, and a second linear layer from 100 dimensions to a scalar RUL prediction with a \texttt{Softplus} output activation and linear scaling. Dropout with rate $r=0.20$ is applied to the final LSTM hidden state and within the fully connected head.

We consider three training approaches: ETO, Decision-Oriented Tuning (DOT), and IEO. All three approaches use the same model architecture and data splits, differing only in their training and model-selection criteria:
\begin{itemize}
\item \textbf{ETO.}
    The prognostic model is trained by minimizing the Mean Squared Error (MSE) over samples from the 70 training engines. After each epoch, the MSE is evaluated on the validation set from 10 validation engines. Training is early-stopped if the validation MSE does not improve for 10 consecutive epochs.
\item \textbf{DOT.}
    Following the decision-oriented tuning strategy of~\citet{Kamariotis-2024-RESS}, the model is still trained using MSE. However, model selection and early stopping are based on the MMC evaluated over the validation set. Training is terminated if the validation MMC does not improve for 10 consecutive epochs.
\item \textbf{IEO.}
    The model is initialized with the ETO-trained parameter and subsequently fine-tuned using the central-difference procedure developed in Section~\ref{sec:4_2}. As in DOT, model selection and early stopping are based on the validation MMC, with a patience of 10 epochs.
\end{itemize}

For all approaches, we use a batch size of 100. To account for sensitivity to the learning rate, for each subset and each random data split, we train each approach using $\eta \in \{1,2,5,10\}\times10^{-4}$.
The learning rate yielding the best validation criterion is selected for test-set evaluation. The training configurations are summarized in Table~\ref{tab:1}.
\begin{table*}[h]
    \centering
    \caption{Training and model-selection configurations of ETO, DOT, and IEO.}
    \begin{tabular}{c c c c c}
    \toprule
         Approach & \makecell{Training/Validation\\Criterion} & Batch Size & Patience & Learning Rate \\
    \midrule
         ETO & MSE/MSE & 100 & 10 epochs & $\{1,2,5,10\} \times 10^{-4}$ \\
         DOT & MSE/MMC & 100 & 10 epochs & $\{1,2,5,10\} \times 10^{-4}$ \\
         IEO & MMC/MMC & 100 & 10 epochs & $\{1,2,5,10\} \times 10^{-4}$\\
    \bottomrule
    \end{tabular}
    \label{tab:1}
\end{table*}

For the maintenance optimization model, we set $\boldsymbol{\tau}\sim\mathrm{Uniform}\{3,4,5,6,7\}$, $c_h=1$, $c_p=5$, and $H=150$. Our sensitivity analysis in~\ref{appendix-sensitivity} demonstrates that the impact of $H$ is marginal. The feasible decision sets are
\[
    \mathcal{Z}_o=\{0,1,\ldots,99\},\qquad
    \mathcal{Z}_d=\{0,1,\ldots,99\},\qquad
    \mathcal{Z}_r=\{5,10,\ldots,100\}.
\]
Under the nominal setting, we set $c_f=5c_p$~\citep{Drent-2023-MSOM}. In Section~\ref{sec:5_3}, we vary the ratio $c_f/c_p$ to represent different levels of aversion to unplanned failure.

\subsection{Nominal case}\label{sec:5_2}
\begin{figure*}
    \centering
    \includegraphics[width=\linewidth]{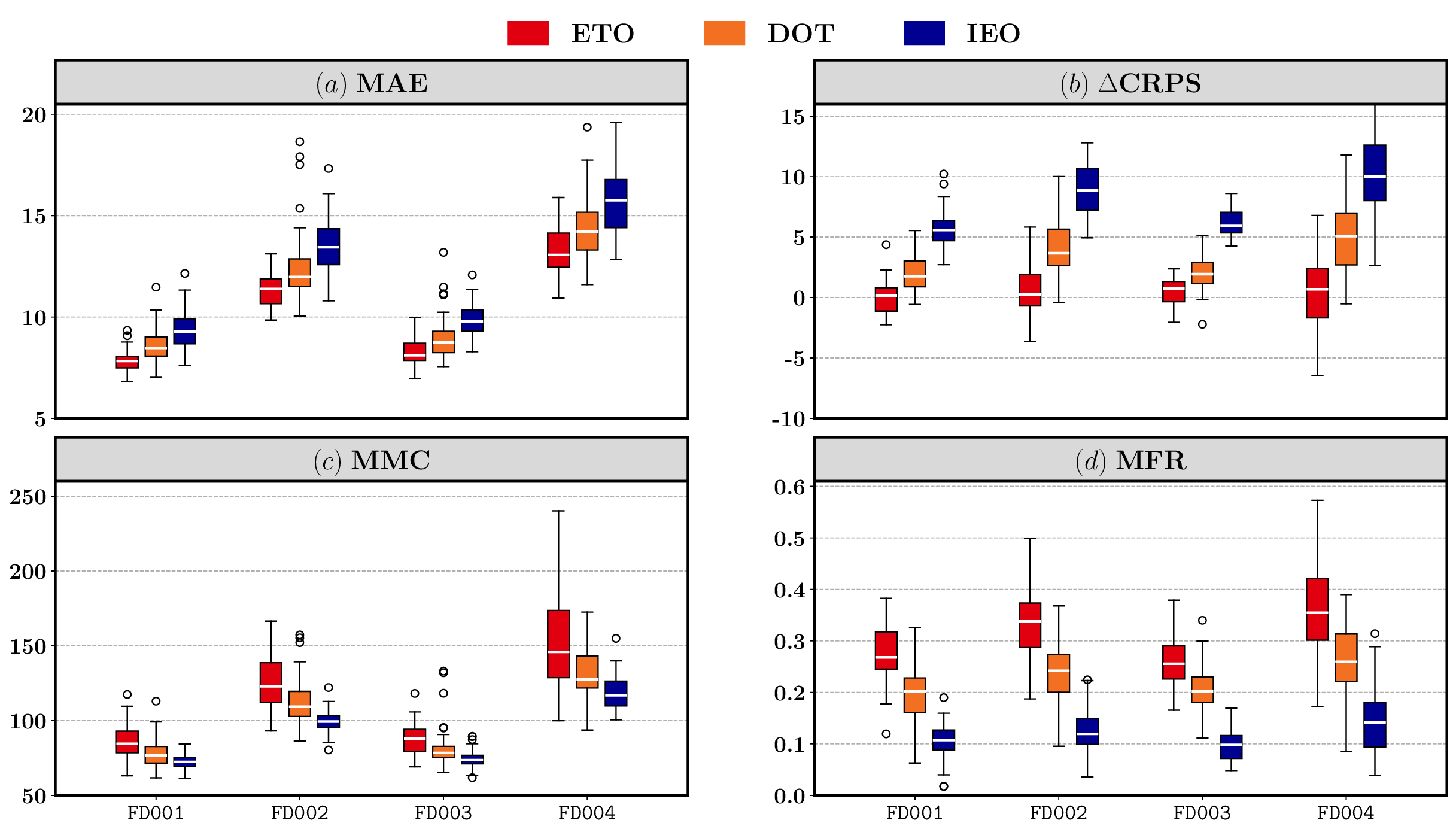}
    \caption{
    Distributions of test-set metrics over 50 random data splits under the nominal setting $c_f/c_p=5$. Panels report (a) MAE, (b) $\Delta$CRPS, (c) MMC, and (d) MFR for ETO, DOT, and IEO across the four CMAPSS subsets. Each box spans the inter-quartile range, the horizontal line denotes the median, and the whiskers extend to the most extreme observations within $1.5$ inter-quartile ranges of the lower and upper quartiles; observations beyond the whiskers are shown individually as outliers.
    }
    \label{fig:case1}
\end{figure*}

In the nominal setting, we set $c_f=5c_p$. To account for variability induced by the data split, we repeat the experiment 50 times for each CMAPSS subset, each time randomly partitioning the 100 engines into training, validation, and test sets according to the $70/10/20$ ratio. The resulting distributions of the test-set metrics across the 50 runs are shown in Figure~\ref{fig:case1}, while the corresponding summary statistics are reported in Table~\ref{tab:2}.

\begin{table*}[h]
\centering
\caption{Test-set performance under the nominal setting $c_f/c_p=5$. Entries report mean $\pm$ standard deviation across 50 random data splits.}\label{tab:2}
\small
\begin{tabular}{ccccccc}
\toprule
Subset & Approach & MAE & CRPS & $\Delta$CRPS & MMC & MFR \\

\midrule
\multirow[c]{3}{*}{\texttt{FD001}}
& ETO & $\mathbf{7.86 \pm 0.50}$ & $\mathbf{5.96 \pm 0.47}$ & $-0.01 \pm 1.30$ & $86.12 \pm 11.17$ & $0.28 \pm 0.05$ \\
& DOT & $8.62 \pm 0.92$ & $6.25 \pm 0.59$ & $1.91 \pm 1.39$ & $77.90 \pm 9.85$ & $0.20 \pm 0.05$ \\
& IEO & $9.36 \pm 0.90$ & $7.69 \pm 0.92$ & $5.70 \pm 1.43$ & $\mathbf{72.29 \pm 5.01}$ & $\mathbf{0.11 \pm 0.03}$ \\
\midrule
\multirow[c]{3}{*}{\texttt{FD002}}
& ETO & $\mathbf{11.31 \pm 0.85}$ & $\mathbf{9.08 \pm 0.85}$ & $0.48 \pm 2.06$ & $125.49 \pm 17.67$ & $0.34 \pm 0.07$ \\
& DOT & $12.46 \pm 1.79$ & $10.10 \pm 1.65$ & $4.26 \pm 2.34$ & $113.01 \pm 16.57$ & $0.24 \pm 0.06$ \\
& IEO & $13.48 \pm 1.43$ & $11.51 \pm 1.48$ & $8.75 \pm 2.16$ & $\mathbf{99.70 \pm 7.85}$ & $\mathbf{0.13 \pm 0.04}$ \\
\midrule
\multirow[c]{3}{*}{\texttt{FD003}}
& ETO & $\mathbf{8.23 \pm 0.68}$ & $\mathbf{6.29 \pm 0.57}$ & $0.50 \pm 1.17$ & $87.25 \pm 10.71$ & $0.26 \pm 0.05$ \\
& DOT & $8.98 \pm 1.10$ & $6.64 \pm 0.88$ & $2.02 \pm 1.36$ & $81.69 \pm 13.31$ & $0.21 \pm 0.05$ \\
& IEO & $9.82 \pm 0.82$ & $8.11 \pm 0.77$ & $6.21 \pm 1.11$ & $\mathbf{74.30 \pm 5.98}$ & $\mathbf{0.10 \pm 0.03}$ \\
\midrule
\multirow[c]{3}{*}{\texttt{FD004}}
& ETO & $\mathbf{13.27 \pm 1.21}$ & $\mathbf{11.00 \pm 1.19}$ & $0.57 \pm 3.04$ & $153.00 \pm 33.74$ & $0.36 \pm 0.09$ \\
& DOT & $14.31 \pm 1.54$ & $11.91 \pm 1.47$ & $5.07 \pm 2.92$ & $131.29 \pm 17.64$ & $0.26 \pm 0.07$ \\
& IEO & $15.71 \pm 1.61$ & $13.69 \pm 1.62$ & $10.15 \pm 3.05$ & $\mathbf{118.32 \pm 11.47}$ & $\mathbf{0.14 \pm 0.07}$ \\
\bottomrule
\end{tabular}
\end{table*}

Figure~\ref{fig:case1}(c) shows that IEO achieves the lowest MMC across all subsets. Compared with ETO, IEO reduces MMC by 16.1\%, 20.6\%, 14.8\%, and 22.7\% on \texttt{FD001}--\texttt{FD004}, respectively. The DOT approach also improves upon ETO, reducing mean MMC by $6.4$--$14.2\%$, which demonstrates the value of using maintenance cost for hyperparameter tuning and model selection as observed by~\citet{Kamariotis-2024-RESS}. Nevertheless, IEO achieves further cost reductions of $7.2$--$11.8\%$ relative to DOT, supporting the benefit of incorporating the maintenance objective into model parameter updates as well as model selection. These improvements also extend to the variability, measured by the standard deviation, of maintenance outcomes. Figure~\ref{fig:case1}(c) shows lower and more concentrated cost distributions for IEO, with the standard deviation of MMC decreasing by $44.2$--$66.0\%$ relative to ETO and $35.0$--$55.1\%$ relative to DOT. Therefore, across the 50 runs, IEO produces both lower average costs and less variation in costs under different data splits. In terms of statistical significance, IEO achieves significantly lower MMC than the baselines under a paired Wilcoxon signed-rank test with $p$-values ($5.73\times 10^{-11}, 1.31\times10^{-4}$) for \texttt{FD001}, ($6.22\times 10^{-14}, 2.95\times10^{-10}$) for \texttt{FD002}, ($3.30\times 10^{-13}, 2.84\times10^{-6}$) for \texttt{FD003}, and ($1.81\times 10^{-12}, 5.64\times10^{-8}$) for \texttt{FD004}, where each pair reports (IEO vs. ETO, IEO vs. DOT).

The cost improvements are accompanied by a tradeoff in prognostic accuracy. Relative to ETO, IEO increases MAE by $18.4$--$19.3\%$ and CRPS by $24.5$--$29.0\%$. However, the derived $\Delta$CRPS reveals that these increased errors have a specific direction. IEO yields positive mean $\Delta$CRPS values between 5.70 and 10.15, exceeding those of both ETO and DOT on every subset and indicating a stronger tendency to underestimate RUL. A similar but more modest directional shift is observed for DOT. This shift is consistent with the lower frequency of failure shown in Figure~\ref{fig:case1}(d): IEO reduces MFR to $10$--$14\%$ relative to ETO ($26$--$36\%$) and DOT ($20$--$26\%$). Under the nominal setting, such a shift is economically reasonable because unplanned downtime carries a higher unit cost than planned downtime. Although premature decommissioning also incurs a cost in predictive maintenance, the simultaneous reductions in MMC and MFR suggest that IEO achieves a more favorable balance between forfeited useful life and unplanned downtime.

The comparison between metrics further illustrates the mismatch between prognostic accuracy and decision quality discussed in Section~\ref{sec:3}. 
DOT incurs only modest increases in MAE and CRPS relative to ETO, yet reduces MMC. Meanwhile, the $\Delta$CRPS of DOT remarkably increases, revealing that conventional prediction accuracy can accompany lower maintenance cost and a stronger tendency toward RUL underestimation.
This observation is consistent with Example~\ref{exam1} and the analysis in Section~\ref{sec:3}, where estimates with comparable accuracy can prescribe different maintenance decisions and incur different costs. More broadly, MAE and CRPS agree that IEO has larger prediction errors, whereas MMC and MFR agree that it yields better maintenance outcomes under the nominal setting. These results provide empirical evidence that prognostic accuracy (MAE and CRPS) alone is insufficient to evaluate a prognosis-then-maintenance policy. Overall, these findings support the motivation for IEO: to use maintenance cost directly as the training objective, rather than as a post-hoc evaluation metric or an indirect model-selection criterion.

\subsection{Varying levels of risk aversion}\label{sec:5_3}
In this section, we vary the level of aversion to unplanned downtime, i.e., the $c_f/c_p$ ratio, in $\{2,5,10,20\}$ to further illustrate the inconsistency between a conventional accuracy metric and the decision objective, and to dissect the value of the IEO framework in industrial practice.

\begin{figure*}[h]
    \centering
    \includegraphics[width=\linewidth]{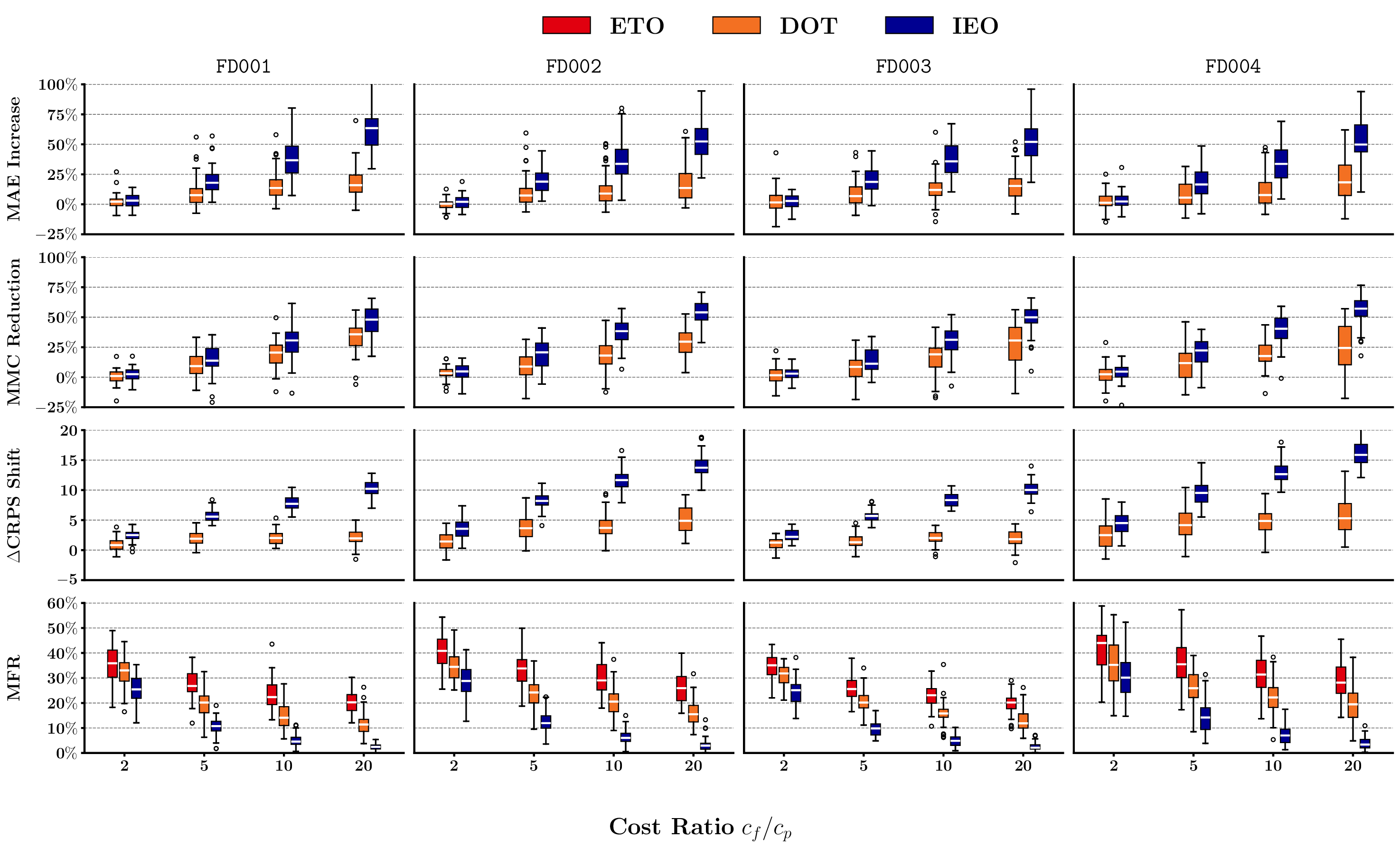}
    \caption{
    Effects of the cost ratio $c_f/c_p$ on prognostic accuracy and maintenance performance across the four CMAPSS subsets. Relative to ETO, the panels report MAE increase, MMC reduction, and $\Delta$CRPS shift for DOT and IEO, together with the MFR of all three approaches. Detailed summary statistics are reported in Table~\ref{tab:case2_more}.
    }
    \label{fig:case2}
\end{figure*}

To illustrate how the inconsistency changes with the ratio $c_f / c_p$, we use derived metrics in this section. Specifically, for each random split, we use the ETO metric as the reference and compute the corresponding changes for DOT and IEO relative to ETO.
Figure~\ref{fig:case2} illustrates the MAE increase (\%), MMC reduction (\%), and change in $\Delta\mathrm{CRPS}$ toward underestimation for DOT and IEO relative to ETO across different ratios. Additionally, Table~\ref{tab:time} reports the average training time (seconds) of applying each approach with four learning rates. The reported IEO training time includes the ETO warm-start time.

\begin{table*}[t]
\centering
\caption{
Effects of the cost ratio $c_f/c_p$ on prognostic and maintenance performance across the four CMAPSS subsets. Entries report mean $\pm$ standard deviation across 50 random data splits. MAE Increase and MMC Reduction are relative percentage changes with respect to ETO, $\Delta$CRPS Shift is the difference relative to ETO, and MFR is reported as a percentage.
}
\label{tab:case2_more}
\resizebox{\textwidth}{!}{
\begin{tabular}{cc cc cc cc ccc}
\toprule
& &
\multicolumn{2}{c}{\textbf{MAE Increase (\%)}} &
\multicolumn{2}{c}{\textbf{MMC Reduction (\%)}} &
\multicolumn{2}{c}{\textbf{$\Delta$CRPS Shift}} &
\multicolumn{3}{c}{\textbf{MFR (\%)}} \\
\cmidrule(lr){3-4}
\cmidrule(lr){5-6}
\cmidrule(lr){7-8}
\cmidrule(lr){9-11}
\textbf{Subset} &
$\boldsymbol{c_f/c_p}$ &
\textbf{DOT} &
\textbf{IEO} &
\textbf{DOT} &
\textbf{IEO} &
\textbf{DOT} &
\textbf{IEO} &
\textbf{ETO} &
\textbf{DOT} &
\textbf{IEO} \\
\midrule
\multirow{4}{*}{\texttt{FD001}} & 2 & $2.6 \pm 6.0$ & $3.1 \pm 5.4$ & $0.7 \pm 5.7$ & $2.1 \pm 5.6$ & $0.9 \pm 1.0$ & $2.4 \pm 0.8$ & $35.7 \pm 6.6$ & $32.3 \pm 6.0$ & $25.8 \pm 5.7$ \\
 & 5 & $9.9 \pm 12.2$ & $19.4 \pm 11.6$ & $8.8 \pm 12.2$ & $14.8 \pm 11.6$ & $1.9 \pm 1.1$ & $5.7 \pm 1.0$ & $27.6 \pm 5.4$ & $20.1 \pm 5.3$ & $10.6 \pm 3.4$ \\
 & 10 & $15.9 \pm 13.2$ & $37.4 \pm 15.1$ & $19.3 \pm 10.8$ & $29.3 \pm 13.7$ & $2.1 \pm 1.2$ & $7.9 \pm 1.2$ & $23.2 \pm 5.9$ & $14.7 \pm 5.0$ & $5.0 \pm 2.5$ \\
 & 20 & $18.1 \pm 12.5$ & $61.0 \pm 16.6$ & $31.4 \pm 18.1$ & $47.0 \pm 12.1$ & $2.1 \pm 1.4$ & $10.3 \pm 1.2$ & $20.3 \pm 4.7$ & $11.5 \pm 4.2$ & $2.4 \pm 1.3$ \\
\midrule
\multirow{4}{*}{\texttt{FD002}} & 2 & $-0.2 \pm 4.7$ & $1.7 \pm 5.6$ & $3.7 \pm 5.1$ & $4.8 \pm 6.0$ & $1.5 \pm 1.5$ & $3.5 \pm 1.5$ & $40.6 \pm 6.7$ & $35.3 \pm 6.3$ & $29.1 \pm 6.3$ \\
 & 5 & $10.2 \pm 13.0$ & $19.4 \pm 10.9$ & $9.2 \pm 12.0$ & $19.3 \pm 11.2$ & $3.8 \pm 2.0$ & $8.3 \pm 1.4$ & $33.7 \pm 6.5$ & $23.8 \pm 5.9$ & $12.8 \pm 4.4$ \\
 & 10 & $12.3 \pm 14.2$ & $36.2 \pm 16.2$ & $17.5 \pm 14.3$ & $37.1 \pm 11.0$ & $4.1 \pm 2.1$ & $11.5 \pm 1.7$ & $30.1 \pm 6.3$ & $20.7 \pm 5.9$ & $6.4 \pm 3.1$ \\
 & 20 & $16.9 \pm 14.9$ & $53.8 \pm 18.4$ & $28.2 \pm 15.0$ & $53.9 \pm 9.2$ & $5.1 \pm 2.1$ & $14.1 \pm 2.0$ & $26.3 \pm 5.8$ & $16.3 \pm 5.1$ & $3.4 \pm 2.5$ \\
\midrule
\multirow{4}{*}{\texttt{FD003}} & 2 & $2.7 \pm 10.0$ & $2.2 \pm 5.6$ & $1.0 \pm 8.9$ & $3.0 \pm 5.9$ & $1.1 \pm 1.0$ & $2.5 \pm 0.9$ & $34.3 \pm 5.4$ & $30.6 \pm 4.7$ & $24.5 \pm 5.1$ \\
 & 5 & $9.2 \pm 11.0$ & $19.7 \pm 9.8$ & $5.8 \pm 13.6$ & $13.9 \pm 10.1$ & $1.5 \pm 1.2$ & $5.7 \pm 1.0$ & $26.2 \pm 4.9$ & $20.5 \pm 4.8$ & $9.8 \pm 2.9$ \\
 & 10 & $13.2 \pm 12.5$ & $37.6 \pm 13.8$ & $15.6 \pm 17.0$ & $30.9 \pm 11.4$ & $2.0 \pm 1.2$ & $8.4 \pm 1.2$ & $23.2 \pm 4.4$ & $16.0 \pm 4.6$ & $4.8 \pm 2.3$ \\
 & 20 & $16.4 \pm 12.8$ & $53.1 \pm 16.2$ & $25.3 \pm 22.2$ & $48.4 \pm 11.5$ & $1.9 \pm 1.4$ & $10.1 \pm 1.4$ & $19.9 \pm 4.2$ & $12.8 \pm 4.4$ & $2.6 \pm 1.9$ \\
\midrule
\multirow{4}{*}{\texttt{FD004}} & 2 & $2.6 \pm 7.7$ & $3.1 \pm 6.7$ & $1.8 \pm 8.4$ & $4.5 \pm 7.1$ & $2.5 \pm 2.4$ & $4.4 \pm 1.6$ & $41.9 \pm 8.7$ & $35.6 \pm 9.3$ & $30.4 \pm 8.5$ \\
 & 5 & $8.3 \pm 11.7$ & $19.1 \pm 14.4$ & $11.8 \pm 14.2$ & $20.2 \pm 13.3$ & $4.5 \pm 2.7$ & $9.6 \pm 2.0$ & $36.0 \pm 9.4$ & $26.1 \pm 6.8$ & $14.5 \pm 6.5$ \\
 & 10 & $11.4 \pm 14.0$ & $34.4 \pm 15.9$ & $18.7 \pm 11.3$ & $39.2 \pm 12.7$ & $4.8 \pm 2.1$ & $13.0 \pm 1.9$ & $31.5 \pm 7.3$ & $22.2 \pm 6.4$ & $7.5 \pm 4.0$ \\
 & 20 & $20.0 \pm 16.7$ & $52.2 \pm 18.4$ & $24.0 \pm 21.2$ & $55.8 \pm 12.6$ & $5.8 \pm 3.0$ & $16.2 \pm 2.2$ & $28.9 \pm 7.9$ & $19.2 \pm 7.0$ & $3.8 \pm 2.4$ \\
\bottomrule
\end{tabular}
}
\end{table*}

\begin{table}[h]
\centering
\caption{Average total training time per random data split over all four candidate learning rates and all cost ratios. The reported time for IEO includes its ETO warm-start.}\label{tab:time}
\begin{tabular}{ccccccc}
\toprule
Subset         & ETO & DOT & IEO \\
\midrule
\texttt{FD001} & 69 &26 &159 \\
\texttt{FD002} & 81 &31 &173 \\
\texttt{FD003} & 73 &26 &162 \\
\texttt{FD004} & 63 &31 &154 \\
\bottomrule
\end{tabular}
\end{table}

Figure~\ref{fig:case2} shows that increasing the ratio sharpens the inconsistency between prognostic accuracy and maintenance performance. As $c_f/c_p$ increases, IEO generally incurs larger increases in MAE while achieving greater MMC reductions relative to ETO. At $c_f/c_p=20$, the mean MMC reductions are $47.0$--$55.8\%$ across the four subsets, accompanied by mean MAE increases of $52.2$--$61.0\%$. The increasingly positive shift in $\Delta$CRPS and the declining MFR explain the direction of this tradeoff: the predictions move further toward RUL underestimation, consistent with fewer cases in which planned decommissioning is scheduled after the realized failure time. A larger ratio makes such late decommissioning more costly, whereas MSE continues to penalize overestimation and underestimation symmetrically. Thus, the economic consequences of prediction errors become increasingly important relative to their magnitude, reinforcing the distinction between estimation risk and decision risk in Section~\ref{sec:3}. DOT exhibits a more modest adjustment: its shift in $\Delta$CRPS changes only marginally beyond $c_f/c_p=5$. This pattern suggests that selecting a model along an MSE-based training trajectory provides less scope to adapt predictions to the maintenance objective than directly updating model parameters using that objective. The training times in Table~\ref{tab:time} provide further evidence consistent with this interpretation: DOT requires only $26$--$31$ seconds, compared with $63$--$81$ seconds for ETO. Together with DOT's improved maintenance outcomes, its shorter training time is consistent with earlier stopping under the MMC criterion and suggests that continued training for predictive accuracy can become counterproductive for decision quality.

These results suggest that IEO is particularly useful when an imperfect prognostic model must support maintenance decisions with substantial penalties for unplanned downtime. For applications with limited run-to-failure data, an important implication is that better maintenance decisions need not require a more expressive prognostic model: the improvements reported here are obtained with the same LSTM architecture and the same dataset. The case for IEO becomes stronger at high $c_f/c_p$ ratios, where its greater cost savings accompany substantially lower failure rates; at $c_f/c_p=20$, its mean MFR is $2.4$--$3.8\%$, compared with $11.5$--$19.2\%$ for DOT. 
IEO also improves on the gains achieved by cost-based model selection. On \texttt{FD002} under the nominal setting, DOT reduces mean MMC by $9.9\%$ relative to ETO, whereas IEO achieves a $20.6\%$ reduction as reported in Table~\ref{tab:2}. However, Figure~\ref{fig:case2} also shows that the cost gains are small on \texttt{FD001} and \texttt{FD002} at $c_f/c_p=2$, so the benefit of IEO depends on the maintenance setting. Its additional training cost should therefore be considered alongside the expected savings: Table~\ref{tab:time} reports $154$--$173$ seconds, including its ETO warm-start, approximately 2.1--2.4 times ETO's training time on our hardware. In practice, the choice of PdM framework should be guided by maintenance costs and failure rates evaluated on held-out engines under the specified $c_f/c_p$ ratio, with particular attention to whether IEO provides sufficient improvement over DOT to justify the additional training.
\section{Concluding Remarks}\label{sec:6}
This paper studies the consistency between prognostic learning and maintenance decision-making in ML-based data-driven PdM. We demonstrate the potential suboptimality of the conventional ETO framework for maintenance decision-making under inevitable estimation error. To mitigate the impact of model misspecification, we propose an IEO framework that uses maintenance cost as the learning objective to fine-tune an accuracy-oriented prognostic model. We establish the consistency of this loss and derive finite-sample guarantees for out-of-sample maintenance cost under independent sampling and overlapping sliding windows. For practical implementation, we develop a learning procedure combining MC dropout, central-difference gradient approximation, and an ETO warm-start to accommodate discrete maintenance decisions.

Experiments on the four CMAPSS subsets demonstrate that improved maintenance outcomes do not necessarily require improved prognostic accuracy. Under the nominal setting, IEO reduces mean maintenance cost by $14.8$--$22.7\%$ relative to ETO and by $7.2$--$11.8\%$ relative to DOT, while also reducing failure rates and cost variation across random data splits. Increasing the ratio sharpens the observed mismatch between accuracy and decision quality: IEO accepts larger prediction errors and shifts predictions further toward RUL underestimation, accompanied by greater cost savings and fewer failures. DOT also improves maintenance outcomes, but its more modest adjustment illustrates the additional value of incorporating maintenance costs into parameter updates. These findings support considering IEO under substantial unplanned-downtime penalties when its additional gains over cost-based model selection justify the extra training cost.

The benefits of IEO come with additional computational expense during training because model updates require solving maintenance optimization problems. Although the warm-start and gradient approximation make this procedure tractable, the reported training time remains approximately $2.1$--$2.4$ times that of ETO, including the warm-start. The practical choice should therefore consider the maintenance improvement over both ETO and DOT alongside the available training resources, especially in settings with low ratios where the benefit can be small. Future work could investigate other prognostic architectures and risk objectives beyond expected maintenance cost, and extend IEO to multi-unit systems with inventory and crew constraints or joint production-maintenance planning. Multistage decision-making and decision-dependent uncertainty represent further directions for future work. On the theoretical side, extending the guarantees to more general temporal dependence and heterogeneous data would strengthen the applicability of IEO beyond the sampling assumptions considered here.





\bibliographystyle{cas-model2-names}
\bibliography{main}

\appendix
\newpage
\section{Omitted Proofs}\label{appendix-proofs}
\subsection{Proof of Proposition~\ref{prop1}}\label{proofofProposition1}
\begin{proofof}{Proposition}{prop1}
First, we prove Assumption~\ref{A2} implies that $g$ is non-constant. Suppose that, to the contrary, there exists $\bz_0 \in \mathcal{Z}$ such that
\begin{equation*}
    g(P) = \bz_0, \quad \forall P \in \operatorname{ri}(\mathcal{P}_{\Xi}),
\end{equation*}
where $\operatorname{ri}(A)$ denotes the relative interior of a set $A$. Optimality of $\bz_0$ implies
\begin{equation*}
    (C_{\bz} - C_{\bz_0})^\top P \geq 0, \quad \forall \bz \in \mathcal{Z}, \quad \forall P \in \operatorname{ri}(\mathcal{P}_{\Xi}).
\end{equation*}
For each $h$, let $P$ approach the simplex vertex assigning probability one to $\xi^h$. By continuity of the preceding linear expression,
\begin{equation*}
    c(\bz; \xi^h) - c(\bz_0; \xi^h) \geq 0, \quad \forall \bz \in \mathcal{Z}, \quad h = 1,\dots, H.
\end{equation*}
Thus $C_{\bz} - C_{\bz_0}$ is component-wise nonnegative for every feasible decision $\bz$.

By Assumption~\ref{A2}, there exist decisions $\bz_\star, \bz_\dagger$ with 
\begin{equation*}
    C_{\bz_\star} \neq C_{\bz_\dagger},
\end{equation*}
and distributions
\begin{equation*}
    P_\star \in S_{\bz_\star} \cap \operatorname{ri}(\mathcal{P}_\Xi), \quad P_\dagger \in S_{\bz_\dagger} \cap \operatorname{ri}(\mathcal{P}_\Xi).
\end{equation*}
Both $\bz_\star$ and $\bz_0$ are optimal under $P_\star$, so 
\begin{equation*}
    (C_{\bz_\star} - C_{\bz_0})^\top P_\star = 0.
\end{equation*}
Every coordinate of $P^\star$ is strictly positive, and every coordinate of $C_{\bz_\star} - C_{\bz_0}$ is nonnegative. Their inner product can therefore vanish only if
\begin{equation*}
    C_{\bz_\star} = C_{\bz_0}.
\end{equation*}
The same argument at $P_\dagger$ gives
\begin{equation*}
    C_{\bz_\dagger} = C_{\bz_0},
\end{equation*}
contradicting $C_{\bz_\star} \neq C_{\bz_\dagger} $. Hence there exist $P_a, P_b \in \operatorname{ri}(\mathcal{P}_\Xi)$ such that
\begin{equation*}
    g(P_a) \neq g(P_b).
\end{equation*}

Next, we construct arbitrarily close distributions that induce different decisions. Define
\begin{equation*}
v \coloneqq P_b-P_a, \quad P(t) := P_a+tv, \quad t \in [0,1]. 
\end{equation*}
Since $P_a$ and $P_b$ assign strictly positive probability to every support point,
\begin{equation*}
\underline{p} \coloneqq  \min_{h=1,\dots,H} \min\{P_a(\xi^h),P_b(\xi^h)\} >0.
\end{equation*}
Therefore,
\begin{equation*}
P(t)(\xi^h) \geq \underline{p}, \quad \forall t \in [0,1], \quad h=1,\dots,H. 
\end{equation*}
For any integer $n \geq 1$, consider the distributions
\begin{equation*}
P(0),P(1/n),\dots,P(1).
\end{equation*}
Because $g(P(0)) \neq g(P(1))$, at least one pair of adjacent distributions must induce different decisions. Hence, there exists $j_n \in \{0,\dots,n-1\}$ such that, defining
\begin{equation*}
P_n \coloneqq P(j_n/n), \quad Q_n \coloneqq P((j_n+1)/n),
\end{equation*}
we have
\begin{equation*}
g(P_n) \neq g(Q_n), \quad Q_n-P_n = \frac{v}{n}.
\end{equation*}
In particular, $P_n$ and $Q_n$ are distinct and belong to $\operatorname{ri}(\mathcal{P}_{\Xi})$.

Since $\mathcal{Z}$ is finite and Assumption~\ref{A1} ensures that it contains at least two distinct decisions, the minimum separation between distinct feasible decisions satisfies
\begin{equation*}
d_{\mathcal{Z}} \coloneqq \min_{\substack{\bz,\bz' \in \mathcal{Z}\\ \bz \neq \bz'}} \|\bz-\bz'\|_1 >0.
\end{equation*}
It follows that
\begin{equation*}
\|g(P_n)-g(Q_n)\|_1 \geq d_{\mathcal{Z}}, \quad \forall n \geq 1,
\end{equation*}
whereas
\begin{equation*}
\|P_n-Q_n\|_{\mathrm{TV}} = \frac{1}{2}\|P_n-Q_n\|_1 = \frac{\|v\|_1}{2n} \longrightarrow 0.
\end{equation*}

Finally, we show that the KL divergences in both directions also converge to zero. For strictly positive probability vectors $P,Q$, write $P_h=P(\xi^h)$ and $Q_h=Q(\xi^h)$. Using $\log u \leq u-1$ for $u>0$, we obtain
\begin{equation*}
\begin{aligned}
\mathbb{D}_{\mathrm{KL}}(P \Vert Q)   &= \sum_{h=1}^{H}P_h\log\frac{P_h}{Q_h}\\
                        &\leq \sum_{h=1}^{H}P_h \left(\frac{P_h}{Q_h}-1\right)\\
                        &= \sum_{h=1}^{H}\frac{(P_h-Q_h)^2}{Q_h},
\end{aligned}
\end{equation*}
where the last equality uses $\sum_{h=1}^{H}P_h=\sum_{h=1}^{H}Q_h=1$. Since every coordinate of $P_n$ and $Q_n$ is at least $\underline{p}$, this gives
\begin{equation*}
\mathbb{D}_{\mathrm{KL}}(P_n \Vert Q_n) \leq \frac{\|P_n-Q_n\|_2^2}{\underline{p}} = \frac{\|v\|_2^2}{\underline{p}\,n^2}.
\end{equation*}
Similarly,
\begin{equation*}
\mathbb{D}_{\mathrm{KL}}(Q_n \Vert P_n) \leq \frac{\|v\|_2^2}{\underline{p}\,n^2}.
\end{equation*}
Consequently,
\begin{equation*}
\max\left\{
\sqrt{\frac{1}{2}\mathbb{D}_{\mathrm{KL}}(P_n\|Q_n)},
\sqrt{\frac{1}{2}\mathbb{D}_{\mathrm{KL}}(Q_n\|P_n)}
\right\}
\leq
\frac{\|v\|_2}{n\sqrt{2\underline{p}}}.
\end{equation*}

Fix any $L_{kl}>0$ and choose $n$ sufficiently large that
\begin{equation*}
\frac{L_{kl}\|v\|_2}{n\sqrt{2\underline{p}}} < d_{\mathcal{Z}}.
\end{equation*}
Then
\begin{equation*}
\begin{aligned}
\|g(P_n)-g(Q_n)\|_1 &\geq d_{\mathcal{Z}}\\
                    &> L_{kl}\min\left\{
\sqrt{\frac{1}{2}\mathbb{D}_{\mathrm{KL}}(P_n\|Q_n)},
\sqrt{\frac{1}{2}\mathbb{D}_{\mathrm{KL}}(Q_n\|P_n)}
\right\}\\
                    &\geq
L_{kl}\|P_n-Q_n\|_{\mathrm{TV}},
\end{aligned}
\end{equation*}
where the last inequality follows from Pinsker's inequality. Taking $P_1=P_n$ and $P_2=Q_n$ proves the claimed inequality for every $L_{kl}>0$. Therefore, $g$ is not globally Lipschitz continuous with respect to total variation distance, which completes the proof.
\qed
\end{proofof}

\subsection{Definition of Natarajan Dimension}\label{sec:B_2}
\begin{definition}[Natarajan Dimension]\label{def:natarajan}
    Consider a policy class $\Pi$ consisting of mappings $\pi:\mathcal{X}\rightarrow\mathcal{Z}$. $\Pi$ Natarajan-shatters a set $\mathcal{X}' \subset \mathcal{X}$ if there exist two policies $\pi_1, \pi_2 \in \Pi$ such that:
    \begin{enumerate}
        \item $\forall x \in \mathcal{X}', \pi_1(x) \neq \pi_2(x)$.
        \item $\forall \mathcal{X}'' \subset \mathcal{X}', \exists\, \pi \in \Pi$, such that $\forall x \in \mathcal{X}'', \pi(x) = \pi_1(x)$, and $\forall x \in \mathcal{X}'\backslash\mathcal{X}'', \pi(x) = \pi_2(x)$.
    \end{enumerate}
    The Natarajan dimension of $\Pi$ is the maximal cardinality of a set $\mathcal{X}'$ that can be shattered by $\Pi$.
\end{definition}

\subsection{Proof of Theorem \ref{theorem-1}}\label{proofoftheorem1}
\begin{proofof}{Theorem}{theorem-1}
For any $N \in \mathbb{N}$, define the growth function of \(\Pi\) by
\begin{equation*}
    G_\Pi(N) \coloneqq \max_{ (x_1,\dots, x_N) \in \mathcal{X}^N } \left|
    \lrc{
    \lra{ \pi_\omega(x_1), \dots, \pi_\omega(x_N) } : \pi_\omega \in \Pi
    }
    \right|.
\end{equation*}
For any fixed set of \(N\) contexts, restricting \(\Pi\) to these contexts gives a multiclass hypothesis class whose label set has cardinality at most \(K\). By Natarajan's lemma~\citep[lemma 29.4]{Shalev-2014-book},
\begin{equation*}
    G_\Pi(N) \leq N^{d_\Pi} K^{2 d_\Pi}.
\end{equation*}
Under Assumption~\ref{A6}, $d_\Pi \leq D_\Pi < \infty$, and therefore
\begin{equation}
    G_\Pi(N) \leq N^{D_\Pi} K^{2 D_\Pi}. \label{theorem1.1}
\end{equation}

By Assumption~\ref{A5}, define the shifted and rescaled maintenance cost
\begin{equation*}
    \bar{c}(\pi_\omega(x);\xi) \coloneqq \frac{c(\pi_\omega(x);\xi) - C_l}{C_u - C_l},
\end{equation*}
such that $\bar{c}(\pi_\omega(x);\xi) \in [0, 1], \forall x, \xi$. Denote 
\begin{equation*}
\begin{aligned}
    \bar{R}(\omega)   &\coloneqq \E_{\P}\lrb{  \bar{c}(\pi_\omega(X);\bxi )  } = \frac{R(\omega) - C_l}{C_u - C_l}, \\
    \tilde{R}_{\mathcal{D}}(\omega) &\coloneqq \frac{1}{n} \sum_{i=1}^{n} \bar{c}( \pi_\omega(x_i); \xi_i ) = \frac{\hat{R}(\omega) - C_l}{ C_u - C_l}, \quad (x_i, \xi_i) \in \mathcal{D}.
\end{aligned}
\end{equation*}
Next, we establish a uniform bound for $\bar{R}(\omega) - \tilde{R}_\mathcal{D}(\omega)$.

Let $\mathcal{D}' = \lrc{\lra{ x_i', \xi_i' }}_{i=1}^{n}$ be an independent sample drawn from $\P^n$, and define
\begin{equation*}
    \tilde{R}_{\mathcal{D}'}(\omega) \coloneqq \frac{1}{n}\sum_{i=1}^{n} \bar{c}(\pi_\omega(x_i');\xi_i').   
\end{equation*}
Fix $\varepsilon > 0$ and consider the event:
\begin{equation*}
    \mathcal{E} \coloneqq \lrc{ \sup_{\omega \in \Omega} \lrb{ \bar{R}(\omega) - \tilde{R}_{\mathcal{D}}(\omega) } > \varepsilon }.
\end{equation*}
For every realization $\mathcal{D} \in \mathcal{E}$, there exists some $\omega_\mathcal{D} \in \Omega$ such that
\begin{equation}
    \bar{R}(\omega_\mathcal{D}) - \tilde{R}_\mathcal{D}(\omega_\mathcal{D}) > \varepsilon. \label{B.1}
\end{equation}
Conditional on $\mathcal{D}$, the random variables
\begin{equation*}
    \bar{c}(\pi_{\omega_\mathcal{D}}(x_i');\xi_i'), \quad i = 1,\dots, n, 
\end{equation*}
are independent, take values in $[0,1]$, and have common expectation $\bar{R}(\omega_\mathcal{D})$. Hoeffding's inequality~\citep[Theorem 2]{Hoeffding} therefore gives
\begin{equation}
    \P_{\mathcal{D}'}\lra{
    \tilde{R}_{\mathcal{D}'}(\omega_\mathcal{D}) - \bar{R}(\omega_\mathcal{D}) \leq -\frac{\varepsilon}{2} \mid \mathcal{D}
    } \leq \exp\lra{-\frac{n \varepsilon^2}{2}}. \label{B.2}
\end{equation}
Suppose $n \varepsilon^2 \geq 2 \log 2 $.~\eqref{B.2} then implies
\begin{equation}
    \P_{\mathcal{D}'}\lra{
    \tilde{R}_{\mathcal{D}'}(\omega_\mathcal{D}) - \bar{R}(\omega_\mathcal{D}) > -\frac{\varepsilon}{2} \mid \mathcal{D}
    } \geq \frac{1}{2}. \label{theorem1.4}
\end{equation}
Combining~\eqref{B.1} and~\eqref{theorem1.4}, whenever $\mathcal{D} \in \mathcal{E}$,
\begin{equation*}
\tilde{R}_{\mathcal{D}'}(\omega_\mathcal{D}) - \tilde{R}_\mathcal{D}(\omega_\mathcal{D}) > \frac{\varepsilon}{2}
\end{equation*}
with conditional probability at least \(1/2\). Consequently,
\begin{equation}
\P_\mathcal{D}(\mathcal{E}) \leq 2 \P_{ \mathcal{D}, \mathcal{D}' } \lra{
\sup_{\omega \in \Omega} \lrb{
\tilde{R}_{\mathcal{D}'}(\omega) - \tilde{R}_\mathcal{D}(\omega) 
} > \frac{\varepsilon}{2}
}.\label{B.0}
\end{equation}
It remains to bound the probability on the right-hand side.

For each $i = 1, \dots, n$, introduce an independent Rademacher random variable $\sigma_i$, satisfying $\sigma_i \sim \mathrm{Uniform}\{-1,1\}$. Since the two observations $(X_i, \bxi_i)$ and $(X_i', \bxi_i')$ are i.i.d. by Assumption~\ref{A3}, their joint distribution is invariant under pairwise exchange. Hence $\tilde{R}_{\mathcal{D}'} - \tilde{R}_{\mathcal{D}}$ has the same distribution as
\begin{equation*}
    \frac{1}{n} \sum_{i=1}^{n} \sigma_i \lrb{
    \bar{c}(\pi_\omega(x_i');\xi_i') - \bar{c}(\pi_\omega(x_i);\xi_i)
    }.
\end{equation*}
Conditional on $\mathcal{D}$ and $\mathcal{D}'$, define
\begin{equation*}
    a_i(\omega) \coloneqq \bar{c}(\pi_\omega(x_i');\xi_i') - \bar{c}(\pi_\omega(x_i);\xi_i).
\end{equation*}
Since \(\bar c\in[0,1]\), $a_i(\omega) \in [-1,1]$. For a fixed \(\omega\), the random variables \(\sigma_i a_i(\omega)\) are independent, have zero mean, and satisfy
\begin{equation*}
    - |a_i(\omega)| \leq \sigma_i a_i(\omega) \leq |a_i(\omega)|.
\end{equation*}
Applying Hoeffding's inequality conditionally on $\mathcal{D},\mathcal{D}'$ yields
\begin{equation}
    \P_\sigma\lra{
    \frac{1}{n}\sum_{i=1}^{n} \sigma_i a_i(\omega) > \frac{\varepsilon}{2} \mid \mathcal{D}, \mathcal{D}'
    } \leq \exp \lra{- \frac{n\varepsilon^2}{8}} \label{B.3}
\end{equation}
Now fix the combined sample
\begin{equation*}
    \mathcal{D}^\star = \lrc{ \lra{x_i, \xi_i}, \lra{x'_i, \xi'_i} }_{i=1}^{n}.
\end{equation*}
For any policy \(\pi_\omega\), all \(2n\) cost values on \(\mathcal{D}^\star\) are determined by the decision vector
\begin{equation*}
\bigl( \pi_\omega(x_1),\ldots,\pi_\omega(x_n), \pi_\omega(x'_1),\ldots,\pi_\omega(x'_n) \bigr).
\end{equation*}
The number of distinct such decision vectors is at most \(G_\Pi(2n)\). Therefore, applying the union bound to~\eqref{B.3},
\begin{equation}
    \P_\sigma\lra{ \sup_{\omega \in \Omega}
    \frac{1}{n}\sum_{i=1}^{n} \sigma_i a_i(\omega) > \frac{\varepsilon}{2} \mid \mathcal{D}, \mathcal{D}'
    } \leq G_\Pi(2n) \exp \lra{- \frac{n\varepsilon^2}{8}}. \label{B.4}
\end{equation}
Taking expectation with respect to \(\mathcal{D},\mathcal{D}'\) and using~\eqref{B.0} gives
\begin{equation}
\P_\mathcal{D} \lra{
\sup_{\omega \in \Omega} \lrb { \bar{R}(\omega) - \tilde{R}_\mathcal{D}(\omega) } > \varepsilon } \leq 
2 G_\Pi(2n) \exp \lra{- \frac{n\varepsilon^2}{8}}. \label{B.5}
\end{equation}
By~\eqref{theorem1.1},
\begin{equation*}
    G_\Pi(2n) \leq (2n)^{D_\Pi} K^{2 D_\Pi},
\end{equation*}
and hence
\begin{equation}
    \log G_\Pi(2n) \leq D_\Pi \log(2n) + 2 D_\Pi \log K. \label{theorem1.2}
\end{equation}
Define
\begin{equation}
    \varepsilon_n \coloneqq D_\Pi \log(2n) + 2 D_\Pi \log K + \log\frac{2}{\delta}, \label{theorem1.3}
\end{equation}
and choose 
\begin{equation*}
    \varepsilon = \sqrt{\frac{8 \varepsilon_n }{n}}.
\end{equation*}
Since \(\delta\in(0,1)\), this choice satisfies the condition required for~\eqref{theorem1.4}. Substitution into~\eqref{B.5} gives
\begin{equation*}
\P_{\mathcal{D}} \lra{
\sup_{\omega \in \Omega} \lrb{ \bar{R}(\omega) - \tilde{R}_\mathcal{D}(\omega) } > \sqrt{\frac{8 \varepsilon_n}{n}}
}    \leq \delta.
\end{equation*}
Therefore, with probability at least \(1-\delta\), simultaneously for all \(\omega\in\Omega\),
\begin{equation*}
    \bar{R}(\omega) \leq \tilde{R}_\mathcal{D}(\omega) + \sqrt{\frac{8 \varepsilon_n }{n}}.
\end{equation*}
Finally, recall the definition of $\bar{R}(\omega)$ and $\tilde{R}_\mathcal{D}(\omega)$,
\begin{equation*}
    R(\omega) \leq \hat{R}(\omega) + C \sqrt{
        \frac{8}{n} \lra{
        D_\Pi \log (2n) + 2 D_\Pi \log K + \log \frac{2}{\delta}
        }
    }.
\end{equation*}
This completes the proof. \qed
\end{proofof}
\subsection{Proof of Corollary~\ref{corollary-1}}\label{profofcoro1}
\begin{proofof}{Corollary}{corollary-1}
Following the same argument in the proof of Theorem~\ref{theorem-1}, for any $\varepsilon > 0$,
\begin{equation}
\P_\mathcal{D} \lra{
\sup_{\omega \in \Omega}
\lrb{
\bar{R}(\omega) - \tilde{R}_\mathcal{D}(\omega)
}
> \varepsilon
}
\leq
2G_\Pi(2n)\exp\lra{-\frac{n\varepsilon^2}{8}}.
\label{C.1}
\end{equation}
By exchanging the roles of the population and empirical risks in the same symmetrization argument, we similarly obtain
\begin{equation}
\P_\mathcal{D} \lra{
\sup_{\omega \in \Omega}
\lrb{
\tilde{R}_\mathcal{D}(\omega) - \bar{R}(\omega)
}
> \varepsilon
}
\leq
2G_\Pi(2n)\exp\lra{-\frac{n\varepsilon^2}{8}}.
\label{C.2}
\end{equation}
Combining~\eqref{C.1} and~\eqref{C.2} by the union bound gives
\begin{equation}
\P_\mathcal{D} \lra{
\sup_{\omega \in \Omega}
\left|
\bar{R}(\omega) - \tilde{R}_\mathcal{D}(\omega)
\right|
> \varepsilon
}
\leq
4G_\Pi(2n)\exp\lra{-\frac{n\varepsilon^2}{8}}.
\label{C.3}
\end{equation}
Recall from~\eqref{theorem1.2} that
\begin{equation*}
\log G_\Pi(2n) \leq D_\Pi \log(2n) + 2D_\Pi \log K.
\end{equation*}
Define
\begin{equation}
\varepsilon_n' \coloneqq D_\Pi \log(2n) + 2D_\Pi \log K + \log\frac{4}{\delta},
\label{C.4}
\end{equation}
and choose
\begin{equation*}
\varepsilon = \sqrt{\frac{8\varepsilon_n'}{n}}.
\end{equation*}
Substituting into~\eqref{C.3} gives
\begin{equation*}
\P_\mathcal{D} \lra{
\sup_{\omega \in \Omega}
\left|
\bar{R}(\omega) - \tilde{R}_\mathcal{D}(\omega)
\right|
>
\sqrt{\frac{8\varepsilon_n'}{n}}
}
\leq \delta.
\end{equation*}
Therefore, with probability at least $1-\delta$, simultaneously for all $\omega \in \Omega$,
\begin{equation}
\left|
R(\omega) - \hat{R}(\omega)
\right|
\leq
C\sqrt{\frac{8\varepsilon_n'}{n}}.
\label{C.5}
\end{equation}
On the event in~\eqref{C.5}, by the definition of $\hat{\omega}_L$,
\begin{equation*}
\hat{R}(\hat{\omega}_L) \leq \hat{R}(\omega^\star).
\end{equation*}
Hence,
\begin{equation*}
\begin{aligned}
R(\hat{\omega}_L) - R(\omega^\star) &=
\lrb{
R(\hat{\omega}_L) - \hat{R}(\hat{\omega}_L)
}
+
\lrb{
\hat{R}(\hat{\omega}_L) - \hat{R}(\omega^\star)
}
+
\lrb{
\hat{R}(\omega^\star) - R(\omega^\star)
} \\
&\leq
\left|
R(\hat{\omega}_L) - \hat{R}(\hat{\omega}_L)
\right|
+
\left|
\hat{R}(\omega^\star) - R(\omega^\star)
\right| \\
&\leq
2C\sqrt{\frac{8\varepsilon_n'}{n}}.
\end{aligned}
\end{equation*}
Substituting the definition of $\varepsilon_n'$ in~\eqref{C.4} yields
\begin{equation*}
R(\hat{\omega}_L) - R(\omega^\star)
\leq
2C \sqrt{
\frac{8}{n}
\lra{
D_\Pi \log(2n) + 2D_\Pi \log K + \log\frac{4}{\delta}
}
}.
\end{equation*}
This completes the proof. \qed
\end{proofof}

\subsection{Proof of Theorem~\ref{theorem-2}}\label{proofoftheorem2}
\begin{proofof}{Theorem}{theorem-2}
For each $r=1,\dots,W$, define the index set
\begin{equation*}
I_r \coloneqq \lrc{r, r+W, r+2W, \dots} \cap \lrc{1,\dots,n},
\end{equation*}
and denote its cardinality by
\begin{equation*}
n_r \coloneqq |I_r|.
\end{equation*}
The collection $\lrc{I_r}_{r=1}^{W}$ forms a partition of $\lrc{1,\dots,n}$, and therefore
\begin{equation}
\sum_{r=1}^{W} n_r = n. \label{D.1}
\end{equation}
Moreover, since the indices within each $I_r$ are separated by at least $W$, Assumption~\ref{A7} implies that
\begin{equation*}
\lrc{\lra{X_i,\bxi_i}:i\in I_r}
\end{equation*}
is a collection of independent observations. Since all observations share the common marginal distribution $\P$, each subsequence indexed by $I_r$ is therefore i.i.d. according to $\P$.

Define the shifted and rescaled maintenance cost as in the proof of Theorem~\ref{theorem-1},
\begin{equation*}
\bar{c}(\pi_\omega(x);\xi)
\coloneqq
\frac{c(\pi_\omega(x);\xi)-C_l}{C_u-C_l}
\in[0,1],
\end{equation*}
and denote
\begin{equation*}
\bar{R}(\omega)
\coloneqq
\E_\P\lrb{\bar{c}(\pi_\omega(X);\bxi)}.
\end{equation*}
For each $r=1,\dots,W$, define the empirical risk evaluated on the $r$th subsequence by
\begin{equation*}
\tilde{R}_r(\omega)
\coloneqq
\frac{1}{n_r}
\sum_{i\in I_r}
\bar{c}(\pi_\omega(x_i);\xi_i).
\end{equation*}
The empirical risk over the entire sliding-window dataset can then be written as
\begin{equation}
\tilde{R}_\mathcal{D}(\omega)
=
\sum_{r=1}^{W}
\frac{n_r}{n}
\tilde{R}_r(\omega).
\label{D.2}
\end{equation}

Since the observations indexed by each $I_r$ are i.i.d., Theorem~\ref{theorem-1} can be applied to each subsequence with sample size $n_r$ and confidence level $\delta/W$. Therefore, for each $r=1,\dots,W$, with probability at least $1-\delta/W$, simultaneously for all $\omega\in\Omega$,
\begin{equation}
\bar{R}(\omega)
\leq
\tilde{R}_r(\omega)
+
\sqrt{
\frac{8}{n_r}
\lra{
D_\Pi\log(2n_r)
+
2D_\Pi\log K
+
\log\frac{2W}{\delta}
}
}.
\label{D.3}
\end{equation}
Applying the union bound over $r=1,\dots,W$, with probability at least $1-\delta$, inequality~\eqref{D.3} holds simultaneously for all $\omega\in\Omega$ and all $r=1,\dots,W$.

Let
\begin{equation*}
n_+ \coloneqq \max_{r=1,\dots,W}n_r
=
\left\lceil\frac{n}{W}\right\rceil,
\end{equation*}
and define
\begin{equation}
\epsilon_{n,W}
\coloneqq
D_\Pi\log(2n_+)
+
2D_\Pi\log K
+
\log\frac{2W}{\delta}.
\label{D.4}
\end{equation}
Since $n_r\leq n_+$ for every $r$,
\begin{equation*}
D_\Pi\log(2n_r)
+
2D_\Pi\log K
+
\log\frac{2W}{\delta}
\leq
\epsilon_{n,W}.
\end{equation*}
Hence, on the same event of probability at least $1-\delta$,
\begin{equation}
\bar{R}(\omega)-\tilde{R}_r(\omega)
\leq
\sqrt{\frac{8\epsilon_{n,W}}{n_r}},
\qquad
\forall r=1,\dots,W,\quad \forall\omega\in\Omega.
\label{D.5}
\end{equation}

Using~\eqref{D.2} and~\eqref{D.5},
\begin{equation*}
\begin{aligned}
\bar{R}(\omega)-\tilde{R}_\mathcal{D}(\omega)
&=
\sum_{r=1}^{W}
\frac{n_r}{n}
\lrb{
\bar{R}(\omega)-\tilde{R}_r(\omega)
} \\
&\leq
\sum_{r=1}^{W}
\frac{n_r}{n}
\sqrt{\frac{8\epsilon_{n,W}}{n_r}} \\
&=
\frac{\sqrt{8\epsilon_{n,W}}}{n}
\sum_{r=1}^{W}\sqrt{n_r}.
\end{aligned}
\end{equation*}
By the Cauchy--Schwarz inequality and~\eqref{D.1},
\begin{equation*}
\sum_{r=1}^{W}\sqrt{n_r}
\leq
\sqrt{
W\sum_{r=1}^{W}n_r
}
=
\sqrt{Wn}.
\end{equation*}
Therefore, with probability at least $1-\delta$, 
\begin{equation}
\bar{R}(\omega)
\leq
\tilde{R}_\mathcal{D}(\omega)
+
\sqrt{
\frac{8W\epsilon_{n,W}}{n}
},
\qquad
\forall\omega\in\Omega.
\label{D.6}
\end{equation}

Finally, recall that
\begin{equation*}
\bar{R}(\omega)
=
\frac{R(\omega)-C_l}{C_u-C_l},
\qquad
\tilde{R}_\mathcal{D}(\omega)
=
\frac{\hat{R}(\omega)-C_l}{C_u-C_l}.
\end{equation*}
Rescaling~\eqref{D.6}, defining $C=C_u-C_l$, and substituting~\eqref{D.4} yield
\begin{equation*}
R(\omega)
\leq
\hat{R}(\omega)
+
C
\sqrt{
\frac{8W}{n}
\lra{
D_\Pi \log \lra{2\left\lceil\frac{n}{W}\right\rceil}
+
2D_\Pi\log K
+
\log\frac{2W}{\delta}
}
}.
\end{equation*}
This completes the proof. \qed
\end{proofof}

\subsection{Finite Natarajan Dimension of the Policy Class}\label{propLSTM}
\begin{proposition}\label{prop:3}
Suppose Assumption~\ref{A4} holds. Consider the LSTM model $m(\cdot;\omega,\mathbf b)$ with parameter space $\Omega\subset\mathbb R^q$, a fixed finite architecture, standard sigmoid and hyperbolic-tangent activations, a ReLU activation, and a softplus output activation. Fix the sliding-window length $W$, the number of sensors $n_s$, the initial hidden and cell states, and the dropout rate $r\in(0,1)$. Let $\mathcal B=\{0,1\}^{d_b}$ and $p_r$ be the dropout-mask space and probability mass function defined in Section~\ref{sec:4_1}.

Suppose scalar predictions are assigned to the $H$ support points in $\Xi$ through fixed RUL intervals $I_1,\dots,I_H$, including endpoint and tail conventions that assign every nonnegative prediction to exactly one interval. Let $m(x;\omega)$ denote the exact dropout-induced RLD defined in Section~\ref{sec:4_1}, obtained by marginalizing $m(x;\omega,\mathbf b)$ over the dropout randomness. Let the optimization model $g$ select a minimizer according to a fixed ordering of $\mathcal Z$ whenever multiple optimal decisions exist.

Then the induced policy class
\begin{equation*}
    \Pi \coloneqq \left\{
    \pi_\omega(\cdot) \coloneqq g\lra{m(\cdot;\omega)} : \mathcal X\rightarrow\mathcal Z \mid \omega\in\Omega 
    \right\}
\end{equation*}
has finite Natarajan dimension, i.e., there exists a constant
$D_\Pi<\infty$ such that
\begin{equation*}
    d_\Pi\leq D_\Pi<\infty.
\end{equation*}
The bound $D_\Pi$ depends only on the fixed model and optimization
specifications and is independent of the sample size $n$.
Consequently, Assumption~\ref{A6} holds for this policy class.
\end{proposition}

\begin{proofof}{Proposition}{prop:3}
Recall that $\mathcal B=\{0,1\}^{d_b}$ is the finite dropout-mask space and that $p_r(\mathbf b)$ denotes the probability mass assigned to $\mathbf b\in\mathcal B$ under dropout rate $r$.
Since the architecture and $r$ are fixed, both $\mathcal B$ and the probabilities $p_r(\mathbf b)$ are fixed independently of $\omega$ and the sample size $n$.

By the definition of the dropout-induced RLD in Section~\ref{sec:4_1}, its $h$th component can be written exactly as
\begin{equation}
    P_h(x;\omega)
    \coloneqq
    \mathbb{P}_{\mathbf b\sim p_r}
    \lrb{
        m(x;\omega,\mathbf b)\in I_h
    }
    =
    \sum_{\mathbf b\in\mathcal B}
    p_r(\mathbf b)
    \mathbb{I}
    \lrb{
        m(x;\omega,\mathbf b)\in I_h
    },
    \qquad h=1,\ldots,H.
    \label{eq:exact_dropout_rld}
\end{equation}
Hence,
\begin{equation*}
    m(x;\omega)
    =
    \bigl(
        P_1(x;\omega),\dots,P_H(x;\omega)
    \bigr)^\top.
\end{equation*}
Thus, although $m(x;\omega)$ corresponds to the limiting RLD obtained from infinitely many stochastic forward passes, its probability vector is an exact finite weighted sum because the dropout-mask space $\mathcal B$ is finite.

Enumerate
$
\mathcal Z
=
\{\boldsymbol z_1,\dots,\boldsymbol z_K\}
$
according to the fixed ordering used to select among multiple
minimizers. Define
\begin{equation*}
    h_\omega(x,k)
    \coloneqq
    \mathbf{1}
    \lrb{
        \pi_\omega(x)=\boldsymbol z_k
    },
    \qquad
    k=1,\dots,K,
\end{equation*}
and the corresponding binary function class
\begin{equation*}
    \mathcal H
    \coloneqq
    \left\{
        (x,k)\mapsto h_\omega(x,k)
        \mid
        \omega\in\Omega
    \right\}.
\end{equation*}
Thus, $h_\omega$ checks whether the policy selects a specified feasible
decision. We first show that $\mathcal H$ has finite VC dimension and
then relate this conclusion to the Natarajan dimension of $\Pi$.

We begin by showing that $h_\omega$ can be evaluated using a uniformly
bounded number of arithmetic operations, exponential operations,
comparisons, and conditional choices. Since the LSTM architecture,
sliding-window length $W$, and initial hidden and cell states are fixed,
each forward pass consists of finitely many affine transformations,
component-wise products, and activation evaluations. The sigmoid and
hyperbolic-tangent activations satisfy
\begin{equation*}
    \sigma(t)
    =
    \frac{1}{1+\exp(-t)},
    \qquad
    \tanh(t)
    =
    \frac{\exp(2t)-1}{\exp(2t)+1}.
\end{equation*}
Both can therefore be evaluated using arithmetic operations and
exponentials, with strictly positive denominators. The ReLU activation
can be evaluated as
\begin{equation*}
    \operatorname{ReLU}(t)
    =
    \begin{cases}
        t, & t>0,\\
        0, & t\leq0,
    \end{cases}
\end{equation*}
using one comparison and a conditional choice. For a fixed mask
$\mathbf b$, dropout requires only multiplication by its mask entries
and, when applicable, the fixed factor $1/(1-r)$. Reusing the same
weights across the $W$ time steps does not increase the number $q$ of
model parameters.

The softplus output requires additional care because logarithms are not
among the operations covered by the result that we will use. Let
$u(x;\omega,\mathbf b)$ denote the scalar value immediately before the
softplus activation, so that
\begin{equation*}
    m(x;\omega,\mathbf b)
    =
    \log
    \lra{
        1+\exp\lra{u(x;\omega,\mathbf b)}
    }.
\end{equation*}
The policy only uses this scalar output to determine its assigned RUL
interval. For every finite interval endpoint $a$, the monotonicity of
the exponential function gives
\begin{equation*}
    m(x;\omega,\mathbf b)\leq a
    \quad\Longleftrightarrow\quad
    1+\exp\lra{u(x;\omega,\mathbf b)}
    \leq
    \exp(a).
\end{equation*}
The same equivalence holds for strict inequalities. Hence, membership of $m(x;\omega,\mathbf b)$ in any interval $I_h$ can be determined using exponentials and comparisons without explicitly evaluating the logarithm. The quantities $\exp(a)$ are fixed constants, and unbounded tail intervals require comparisons only with their finite endpoints.

For each $\mathbf b\in\mathcal B$, the indicator $\mathbf{1} \lrb{ m(x;\omega,\mathbf b)\in I_h }$ can therefore be evaluated using a bounded number of admissible operations. Since $|\mathcal B|=2^{d_b}<\infty$ and $H<\infty$,~\eqref{eq:exact_dropout_rld} shows that the exact RLD $m(x;\omega)$ can be evaluated using finitely many such operations, followed by multiplication by the fixed constants $p_r(\mathbf b)$ and finite summation.

Next, recall that the expected cost of each feasible decision is
\begin{equation*}
    C_{\boldsymbol z_k}^{\top}m(x;\omega)
    =
    \sum_{h=1}^{H}
    c(\boldsymbol z_k;\xi^h)
    P_h(x;\omega),
    \qquad
    k=1,\dots,K.
\end{equation*}
The scenario costs are fixed constants. Since $K<\infty$, the optimization model $g$ can be evaluated by computing these $K$ expected costs and selecting the first decision attaining their minimum according to the fixed ordering of $\mathcal Z$. This requires only finitely many additional arithmetic operations and comparisons. Comparing the selected index with $k$ then yields $h_\omega(x,k)$.

Consequently, there exists an integer $T<\infty$ such that every evaluation of $h_\omega$ requires at most $T$ operations consisting of arithmetic, exponentiation, comparisons, conditional choices, and a binary output. The bound $T$ is uniform over $x$ and $\omega$ and depends only on the fixed model architecture, dropout specification, RUL discretization, and optimization model. In particular, although $T$ may depend on $|\mathcal B|=2^{d_b}$, it is independent of the sample size $n$.

By \citet[Theorem~8.14]{Anthony-1999-book}, a binary function class parameterized by $q$ real numbers and evaluated using at most $T$ such operations has VC dimension bounded by
\begin{equation*}
    \operatorname{VCdim}(\mathcal H)
    \leq
    T^2q
    \bigl(
        q+19\log_2(9q)
    \bigr)
    \coloneqq
    D_\Pi
    <\infty.
\end{equation*}
Here $q\geq1$ for the LSTM under consideration; if $q=0$, the class
contains a single function and the conclusion is immediate. To apply
the cited theorem on Euclidean input and parameter spaces, one can
evaluate the same formulas for all
$x\in\mathbb R^{Wn_s}$ and $\omega\in\mathbb R^q$, and return zero
whenever the decision index $k$ is invalid. These validity checks
require only finitely many additional comparisons, which can be
included in $T$. Restricting the inputs to
$\mathcal X\times\{1,\dots,K\}$ and the parameters to $\Omega$ cannot
increase VC dimension.

It remains to connect this binary bound to the Natarajan dimension.
Suppose that
\begin{equation*}
    \{x_i:i=1,\dots,N\}
\end{equation*}
is Natarajan-shattered by $\Pi$. By
Definition~\ref{def:natarajan}, there exist two policies whose
decisions at $x_i$ are $\boldsymbol z_{k_i}$ and
$\boldsymbol z_{\ell_i}$, respectively, with
$k_i\neq\ell_i$. For every subset
$J\subseteq\{1,\dots,N\}$, there exists $\omega_J\in\Omega$ such that
\begin{equation*}
    \pi_{\omega_J}(x_i)
    =
    \begin{cases}
        \boldsymbol z_{k_i}, & i\in J,\\
        \boldsymbol z_{\ell_i}, & i\notin J.
    \end{cases}
\end{equation*}
Therefore,
\begin{equation*}
    h_{\omega_J}(x_i,k_i)
    =
    \begin{cases}
        1, & i\in J,\\
        0, & i\notin J.
    \end{cases}
\end{equation*}
Thus, $\mathcal H$ realizes every binary labeling of the $N$ points
\begin{equation*}
    \{(x_i,k_i):i=1,\dots,N\}.
\end{equation*}
By the definition of VC dimension,
\begin{equation*}
    N
    \leq
    \operatorname{VCdim}(\mathcal H).
\end{equation*}
It follows that
\begin{equation*}
    d_\Pi
    \leq
    \operatorname{VCdim}(\mathcal H)
    \leq
    D_\Pi
    <\infty.
\end{equation*}
Since $q$ and $T$ depend only on the fixed model and optimization
specifications, the bound $D_\Pi$ is independent of the sample size
$n$. This verifies Assumption~\ref{A6} and completes the proof.
\qed
\end{proofof}
\section{Sensitivity Check}\label{appendix-sensitivity}
We provide the sensitivity analysis results on the choice of support horizon $H$. We consider the candidates $H \in \{100, 125, 150\}$ while other parameters are identical to those in Section~\ref{sec:5_2}, taking $c_f =5 c_p$. The results of MAE, CRPS, and MMC for \texttt{FD001}--\texttt{FD004} are summarized in Table~\ref{tab:sensitivity}. It is clearly shown by the results that the support horizon $H$ has a marginal impact on the prediction accuracy and decision quality. Eventually, we take $H=150$ because it achieves the best performance the most times across different subsets and methods. Note the results in Table~\ref{tab:sensitivity} differ from the results presented in Table~\ref{tab:2} because the experiments for sensitivity analysis are individually executed.

\begin{table*}[h]
\centering
\caption{Sensitivity analysis of $H$. Results are reported as mean $\pm$ standard deviation over 50 runs. Bold values indicate the best performance.}
\label{tab:sensitivity}
\resizebox{\textwidth}{!}{%
\begin{tabular}{llccccccccc}
\toprule
\multirow{2}{*}{Dataset} & \multirow{2}{*}{Method} & \multicolumn{3}{c}{MAE} & \multicolumn{3}{c}{CRPS} & \multicolumn{3}{c}{MMC} \\
\cmidrule(lr){3-5}
\cmidrule(lr){6-8}
\cmidrule(lr){9-11}
& & $H=100$ & $H=125$ & $H=150$ & $H=100$ & $H=125$ & $H=150$ & $H=100$ & $H=125$ & $H=150$ \\
\midrule
\multirow{3}{*}{FD001} & ETO & $7.86 \pm 0.53$ & $\mathbf{7.83 \pm 0.55}$ & $7.90 \pm 0.49$ & $5.99 \pm 0.49$ & $\mathbf{5.95 \pm 0.55}$ & $6.00 \pm 0.47$ & $86.20 \pm 11.94$ & $86.35 \pm 12.78$ & $\mathbf{85.56 \pm 10.99}$ \\
 & DOT & $8.63 \pm 0.85$ & $8.62 \pm 0.74$ & $\mathbf{8.48 \pm 0.81}$ & $6.33 \pm 0.65$ & $6.34 \pm 0.50$ & $\mathbf{6.19 \pm 0.52}$ & $77.80 \pm 9.75$ & $79.17 \pm 8.37$ & $\mathbf{76.97 \pm 9.44}$ \\
 & IEO & $9.38 \pm 0.89$ & $9.39 \pm 0.83$ & $\mathbf{9.30 \pm 0.63}$ & $7.76 \pm 0.89$ & $7.73 \pm 0.84$ & $\mathbf{7.62 \pm 0.61}$ & $72.14 \pm 5.27$ & $\mathbf{72.07 \pm 5.10}$ & $72.16 \pm 5.39$ \\
\midrule
\multirow{3}{*}{FD002} & ETO & $11.36 \pm 0.76$ & $11.44 \pm 0.94$ & $\mathbf{11.28 \pm 0.84}$ & $9.11 \pm 0.78$ & $9.21 \pm 0.93$ & $\mathbf{9.03 \pm 0.85}$ & $129.21 \pm 19.64$ & $127.48 \pm 17.88$ & $\mathbf{124.64 \pm 13.89}$ \\
 & DOT & $12.45 \pm 1.39$ & $\mathbf{12.29 \pm 1.22}$ & $12.40 \pm 1.41$ & $10.07 \pm 1.37$ & $\mathbf{9.96 \pm 1.18}$ & $10.04 \pm 1.39$ & $113.05 \pm 13.70$ & $113.40 \pm 12.92$ & $\mathbf{111.30 \pm 13.21}$ \\
 & IEO & $13.48 \pm 1.30$ & $\mathbf{13.34 \pm 1.42}$ & $13.35 \pm 1.34$ & $11.45 \pm 1.33$ & $\mathbf{11.31 \pm 1.45}$ & $11.33 \pm 1.39$ & $100.12 \pm 7.29$ & $99.68 \pm 8.77$ & $\mathbf{99.50 \pm 7.32}$ \\
\midrule
\multirow{3}{*}{FD003} & ETO & $8.34 \pm 0.79$ & $8.33 \pm 0.97$ & $\mathbf{8.22 \pm 0.66}$ & $6.32 \pm 0.65$ & $6.35 \pm 0.80$ & $\mathbf{6.30 \pm 0.61}$ & $89.37 \pm 11.59$ & $89.93 \pm 12.45$ & $\mathbf{89.00 \pm 9.26}$ \\
 & DOT & $8.90 \pm 1.00$ & $9.01 \pm 1.12$ & $\mathbf{8.82 \pm 0.83}$ & $6.59 \pm 0.80$ & $6.64 \pm 0.84$ & $\mathbf{6.55 \pm 0.71}$ & $\mathbf{81.57 \pm 10.67}$ & $82.57 \pm 13.47$ & $81.71 \pm 9.96$ \\
 & IEO & $9.86 \pm 0.77$ & $9.78 \pm 0.96$ & $\mathbf{9.71 \pm 0.84}$ & $8.14 \pm 0.79$ & $8.09 \pm 0.94$ & $\mathbf{8.04 \pm 0.83}$ & $74.49 \pm 6.16$ & $\mathbf{73.76 \pm 6.05}$ & $74.27 \pm 5.34$ \\
\midrule
\multirow{3}{*}{FD004} & ETO & $\mathbf{13.24 \pm 1.16}$ & $13.43 \pm 1.12$ & $13.43 \pm 1.18$ & $\mathbf{10.95 \pm 1.14}$ & $11.11 \pm 1.09$ & $11.11 \pm 1.12$ & $153.41 \pm 27.14$ & $\mathbf{152.77 \pm 30.96}$ & $153.34 \pm 28.76$ \\
 & DOT & $14.50 \pm 1.57$ & $\mathbf{14.28 \pm 1.22}$ & $14.29 \pm 1.66$ & $12.10 \pm 1.46$ & $\mathbf{11.90 \pm 1.18}$ & $11.92 \pm 1.57$ & $132.39 \pm 19.17$ & $132.11 \pm 19.80$ & $\mathbf{131.61 \pm 19.93}$ \\
 & IEO & $\mathbf{15.66 \pm 1.92}$ & $15.73 \pm 1.70$ & $15.87 \pm 1.69$ & $\mathbf{13.63 \pm 1.93}$ & $13.68 \pm 1.71$ & $13.82 \pm 1.66$ & $\mathbf{116.38 \pm 12.09}$ & $116.72 \pm 10.69$ & $117.36 \pm 12.05$ \\
\bottomrule
\end{tabular}%
}
\end{table*}

\end{document}